%% file: main.tex
\documentclass[letterpaper]{article} 
\usepackage[preprint]{aaai2027}  
\usepackage[hyphens]{url}  
\usepackage{graphicx} 
\usepackage{natbib}  
\usepackage{caption} 
\usepackage{algorithm}
\usepackage{algorithmic}
\usepackage{multirow}
\usepackage{booktabs}

\usepackage{newfloat}
\usepackage{listings}
\DeclareCaptionStyle{ruled}{labelfont=normalfont,labelsep=colon,strut=off} 
\floatstyle{ruled}
\newfloat{listing}{tb}{lst}{}
\floatname{listing}{Listing}

\usepackage{booktabs}

\input{defs}

\title{Communication-Enhanced Tutoring \\ for Efficient Decentralized Multi-Agent Reinforcement Learning}
\author{
    Maciej Wojtala \textsuperscript{\rm 1, \rm 2} \equalcontrib\corresponding \;\;
    Bogusz Stefańczyk \textsuperscript{\rm 2} \equalcontrib \;\; 
    Dominik Jacek Bogucki \textsuperscript{\rm 3, \rm 4} \equalcontrib \;\; 
    Łukasz Lepak \textsuperscript{\rm 5} \;\;  
    Paweł Wawrzyński \textsuperscript{\rm 2}
}
\affiliations{
    \textsuperscript{\rm 1} Univeristy of Warsaw\\
    \textsuperscript{\rm 2} IDEAS Research institute\\
    \textsuperscript{\rm 3} Akces NCBR\\
    \textsuperscript{\rm 4} Institute of Fundamental Technological Research of the Polish Academy of Sciences\\
    \textsuperscript{\rm 5} Warsaw Univeristy of Technology\\
    mw406587@students.mimuw.edu.pl, bogusz.stefanczyk@ideas.edu.pl, dominik.bogucki@akces-ncbr.pl, lukasz.lepak@pw.edu.pl, pawel.wawrzynski@ideas.edu.pl
}

\begin{document}

\maketitle

\begin{abstract}
Centralized Training with Decentralized Execution (CTDE) is the dominant paradigm in multi-agent reinforcement learning (MARL), enabling agents to act independently at test time while leveraging additional information during training. However, the most prominent methods within CTDE, based on value decomposition, are limited in learning efficiency and final performance by partial observability in both training and execution. 
To overcome this limitation, in this work, we propose the~framework of tutoring: In training, the agents share information in their latent space to develop well-informed policies that achieve strong performance. Then, to recover decentralized execution, these policies concurrently adjust to anticipate lack of communication, and they are distilled into counterparts that rely solely on local observations. 
We demonstrate the effectiveness of our approach on Hallway, which, to the best of our knowledge, has not been solved before without test-time communication, SMAC under settings more difficult than the standard ones, and SMACv2.
\end{abstract}

\section{Introduction}

Multi-agent reinforcement learning (MARL) \citep{2024ning+1,2021du+1,2020nguyen+2,2023oroojlooy+1,2023wong+3} offers a powerful framework for controlling large-scale systems by allowing multiple agents to learn collectively while interacting within a shared, dynamic environment. In MARL, agents seek to optimize their action-selection strategies (policies) based on experience derived from their interactions with the environment and other agents. 
MARL enables a broad spectrum of applications, including robotics \citep{2023or+1}, autonomous driving \citep{2023yadav+2,2022schmidt+4,2019althamary+2}, and telecommunications \citep{2022li+6,2021feriani+1}.
However, MARL presents several fundamental challenges: limited and heterogeneous environmental observations across agents, the need for coordination among agents, non-stationarity induced by concurrently evolving policies, and scalability to systems with many agents.
To address these challenges, the dominant framework for cooperative MARL is \emph{Centralized Training and Decentralized Execution} (CTDE) \cite{2016kraemer+1}, in which the learning algorithm has access to the global state during training, whereas each agent relies solely on its local observations at execution time \citep{2018rashid+5,2021wang+4,2019son+4}. 

Sharing information between agents usually improves their performance. MASIA \citep{Guan2022masia} employs a communication mechanism that exchanges agent observations via a Transformer module. CADP \citep{2025zhou+6CADP} adheres to the CTDE paradigm, allowing the agents to share observations via a Transformer encoder during training, while progressively pruning inter-agent connections. 
However, these methods have several drawbacks. MASIA, instead of feeding the Transformer outputs directly to agents, aggregates them into a single vector and learns to reconstruct the global state from this vector using an auxiliary loss. This reconstruction procedure requires the number of parameters to grow with the number of agents.
Both MASIA and CADP apply a Transformer to the agents’ inputs rather than on hidden representations of observation-action histories. Thus, the agents can exchange only inputs from the current time step, not the entire history. In CADP, as pruning proceeds, agents are increasingly constrained to act solely on local observations, thereby increasing the difficulty of the learning problem over time.

Methods such as PTDE \citep{chen2024ptde} or CTDS \citep{zhao2024ctds} employ a teacher-student paradigm to distill knowledge from centralized algorithms using an MSE loss between individual Q-functions conditioned on local and global information.
We claim that distilling the entire local Q-function is more challenging than distilling the best action (i.e., the argmax of the local Q-function).
CTPDE \citep{PEI2025129617} applies the Kullback-Leibler divergence for knowledge distillation in MARL. However, distillation occurs after centralized training, in an offline manner without any mitigation of the imitation gap. Moreover, offline distillation requires the procedure to be performed sequentially by training a centralized model and then distilling a decentralized policy.
Instead, we propose an online distillation method.

The main idea behind our proposed method is to train a well-informed policy under a value-decomposition paradigm using all available information, and concurrently distill it in an online manner into a policy based solely on the agent's observation-action history, leveraging the value-decomposition paradigm's architecture \citep{2022li+7}, applying a simple cross-entropy loss. For well-informed policy training, we use a Transformer-based architecture that processes each agent’s latent representation of the information processed so far and shares it between agents.
In this work, we adhere to the CTDE paradigm but relax its standard training setup: during training, each agent is provided with the latents of all agents, representing all information gathered by all agents during previous steps, thereby enabling access to a richer representation of the environment. The latents are processed by a Transformer-based encoder module \citep{2017vaswani+7}, which keeps the parameter count constant (independent of the number of agents) and enables reasonable scalability. The resulting representations are used to evaluate candidate actions, which are subsequently aggregated during training via a mixing network, such as QMIX \citep{2018rashid+5}. The additional information allows agents to more effectively distinguish between optimal and suboptimal actions, leading to improved performance. To ensure decentralized execution, we distill in an online manner the learned centralized policies into decentralized counterparts that rely solely on local observations. 

\citet{warrington21a} apply an annealed mixture of teacher and student actions to the replay buffer. We employ a similar approach in the value decomposition paradigm by choosing, in exploitation, between the greedy actions of the centralized teacher policy and the decentralized student policy. 
Moreover, to adapt the teacher policy to the absence of communication, we evaluate the subsequent states using mixtures of centralized and decentralized tuples of actions.

\begin{figure*}[t!]
  \centering
  \includegraphics[width=0.97\textwidth]{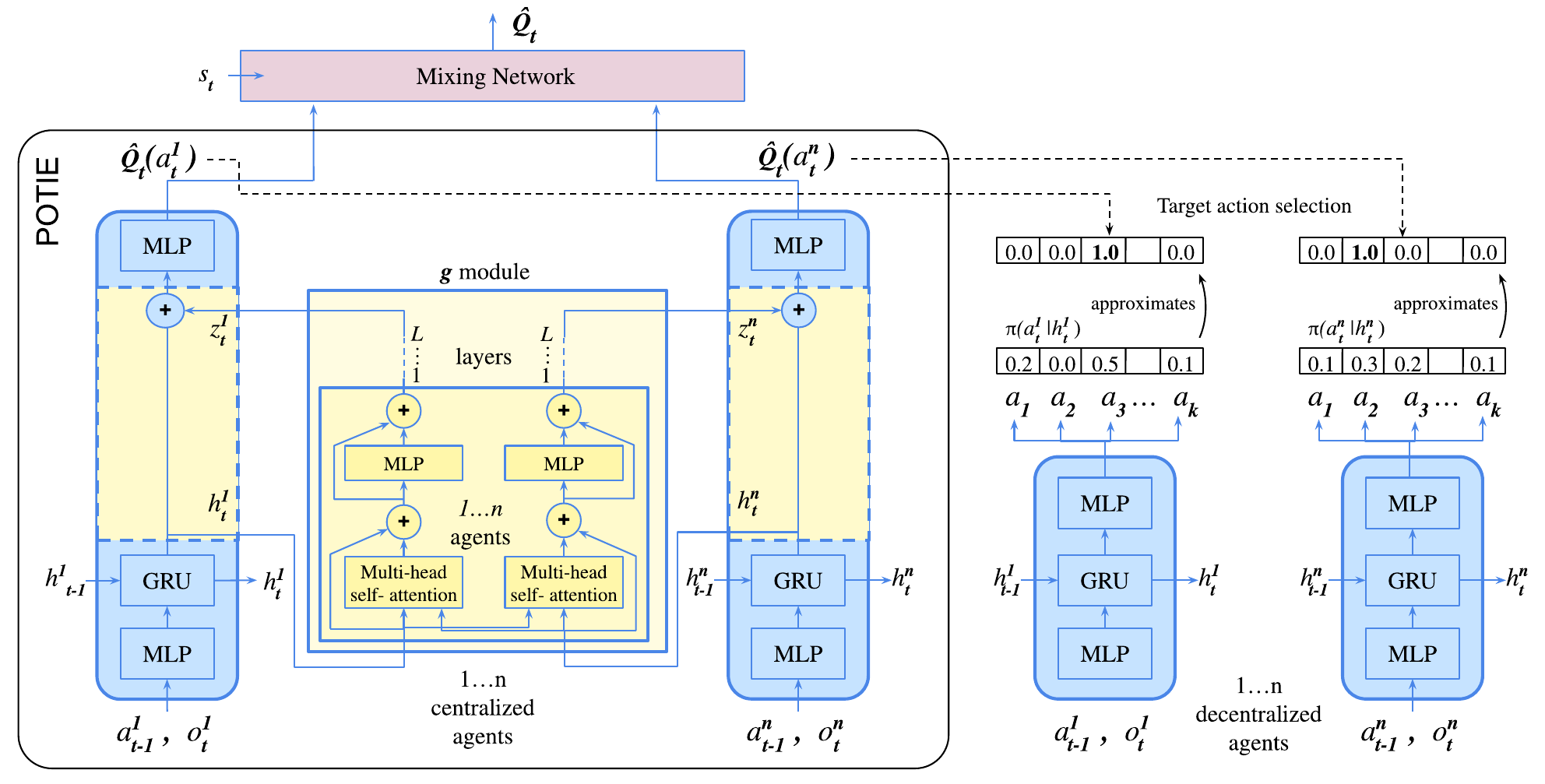}
  \caption{The proposed \ourall{} tutoring mechanism with \ourcom{} as the centralized architecture. On the left, we show \ourcom{}, which consists of a conventional agent architecture used in value-decomposition methods, with an added Transformer encoder module ($g$) that shares latent representations of information gathered by the team of agents during the episode. On the right, we show the distillation part: another network with a conventional agent architecture, but without the g module, that learns through a cross-entropy loss to mimic the policy derived from \ourcom{}.}
  \label{fig:architecture}
\end{figure*}

{\bf The main contributions of this paper} are as follows: 
\begin{itemize}
    \item We introduce the framework of {\it tutoring}, where the agents train well-performing policies with communication. These policies are concurrently adjusted to account for the lack of communication and are distilled into ones that rely on local information only. 
    \item To obtain well-informed policies, we introduce \ourcom{} -- an~effective Transformer-based method for sharing agents' latent representations, enabling the use of all information collected by the entire team of agents during the episode.
    \item To realize tutoring, we present \ourall{} -- an online policy distillation algorithm based on a~cross-entropy loss.
\end{itemize}

\section{Related Work}

\akapit{MARL} \citep{2024ning+1} generalizes single-agent trial-and-error learning \citep{2018sutton+1} to settings where multiple agents simultaneously influence a shared, dynamic environment.
MARL encompasses a wide range of problem formulations that vary along dimensions such as synchronous versus asynchronous decision-making, cooperative versus competitive objectives, and different information structures. In this work, we restrict our attention to the synchronous, fully cooperative setting, where all agents share a common reward signal at each time step. 

\akapit{Value decomposition} 
Value decomposition emerged to address the exponential growth of the joint action-value function with the number of agents \citep{2016kraemer+1}.
Decomposition-based architectures typically assign each agent a local module that processes its observations and previous actions to produce an individual contribution to the joint action-value function \citep{2022li+7}, and a global mixing architecture, used only during training, that combines these contributions into the joint action-value function. An agent’s optimal action is then determined by maximizing its individual contribution to the joint action-value estimate. Several mixing approaches have been proposed in the literature. Value Decomposition Networks (VDN) \citep{2018sunehag+10} aggregate agent contributions through simple summation, while QMIX \citep{2018rashid+5} employs a mixing network that learns a monotonic transformation. QPLEX \citep{2021wang+4} introduces a duplex dueling architecture over agent contributions. We provide an extended literature review of this topic in the Appendix.

\akapit{Inter-agent sharing of observations}
Agents typically perform better with a more comprehensive perception of the environment, which communication mechanisms provide by sharing states or observations. In MAIC \citep{2022yuan+5}, agents explicitly model their teammates and use this information to bias their action-value function. MASIA \citep{Guan2022masia} uses an attention-based communication architecture and a decoder module to reconstruct the global state. CommFormer \citep{hu2024commformer} uses an attention encoder-decoder architecture and learns a communication graph, with the number of parameters independent of the number of agents. We provide an extended literature review of this topic in the Appendix.

\akapit{Policy distillation frameworks} CADP \citep{2025zhou+6CADP} uses gradually pruned communication in training and disables communication during execution. PTDE \citep{chen2024ptde} uses centralized training with the global state fed to the agent network, and then distills that model into a student network offline, in a separate second stage. CTDS \citep{zhao2024ctds} uses a teacher-student framework, where the teacher takes centralized observations as input and distills knowledge to the student using an MSE loss on the individual Q-functions. IGM-DA \citep{hong2022rethinking} uses the DAgger \citep{pmlr-v15-ross11a} algorithm and the Bayesian expected loss \citep{Shao01091989} to distill an expert trained using global information to the student. CTPDE \citep{PEI2025129617} applies the Kullback-Leibler divergence for knowledge distillation in MARL by weighting two components: cross-entropy between the student policy and the argmax of the teacher Q-function, and KL divergence between the student policy and the soft policy obtained from the Q-function.

\akapit{Imitation learning corrections}
Teacher–student distillation is susceptible to the imitation gap. Several studies aim to reduce this gap. \citep{warrington21a} extend the DAgger algorithm \citep{pmlr-v15-ross11a} with an annealed mixture of centralized and decentralized policies and a mixture of centralized and decentralized value functions. \citep{NEURIPS2021_9fc66491} mix the cross-entropy distillation loss between centralized and decentralized policies and the reward-based loss for the decentralized policy. \citep{pmlr-v100-chen20a} use online imitation learning based on the $L_1$ loss.

\section{Method}
\label{section:Method}

In this section, we present the formalism of the problem and describe our proposed solution.

\paragraph{Problem definition.}

In this paper, we discuss the problem of MARL using the formalism of a cooperative Decentralized Partially Observable Markov Decision Process (Dec-POMDP). A Dec-POMDP is defined by a tuple $\langle \agents, \states, \actions, \R, \P, \Omega, \O, \gamma \rangle$, where $\agents = \{ 1, \dots, n \}$ represents the team of agents, $\states$ is the global environmental state space, $\actions = \actions^1 \times \dots \times \actions^n$ is the joint action space, $\R : \states \times \actions \times \states \rightarrow \real$ is the reward function, $\P : \states \times \actions \rightarrow P(\states)$ is the state transition probability function, $\Omega$ is the observation space, $\O: \agents \times \states \rightarrow \Omega$ is the observation function and $\gamma \in [0,1]$ is the discount factor. At time $t=1,2,\dots$ the $i$-th agent makes an observation, $o^i_t = \O(i, s_t)$, of the environment state $s_t \in \states$. Each agent $i$ in time step $t$ maintains an action-observation history $\tau^i_t \in \T_i \equiv (\Omega \times \actions_i)^*$. The policy of an agent is conditioned on its action-observation history $\pi^i: \T_i \rightarrow P(\actions^i)$. The agent performs an action, $a^i_t \in \actions^i$, $a^i_t \sim \pi^i(\cdot|\tau^i_t)$. The joint action $a_t = [(a^1_t), \dots, (a^n_t)]$ affects the next environment state $s_{t+1} \sim \P(\cdot | s_t, a_t)$ and all agents receive the reward $r_t = \R(s_t, a_t, s_{t+1})$.
The agents’ objective is to optimize the joint policy $\pi=[\pi^1, \dots, \pi^n]$ so as to maximize the expected discounted return from each state $s$, the highest sum of discounted rewards: 
$$
    \V^\pi(s) = \mathbb{E}\left(\sum_{k\geq0} \gamma^i r_{t+k} \Big| s_t = s, \text{policy in use} = \pi\right).
$$

\subsection{\ourcom{}: training well-informed policies}

For centralized training, we propose \ourcom{} (POlicy with Transformer-based Information Exchange), shown in Fig.~\ref{fig:architecture}. It follows the value-decomposition paradigm and is based on a Transformer architecture, similarly to MASIA \citep{Guan2022masia} and CADP \citep{2025zhou+6CADP}. However, unlike MASIA and CADP, \ourcom{} processes the entire history of information gathered during the episode.
We start with a standard agent architecture ($f^i$ for the $i$-th agent) that comprises a feedforward part, denoted by MLP in Fig.~\ref{fig:architecture}, a recurrent part, denoted by GRU, and an output feedforward part, again denoted by MLP \citep{2022li+7}. This module produces individual Q-functions ($\est Q^i_t$) from the current observation ($o^i_t$) and the previous action ($a^i_{t-1}$). The outputs of the $f^i$ modules are fed to the mixing network that produces the value of the action-value function, $\est Q_t$.
We divide the $f^i$ module after its recurrent part and feed its latent state $h^i_t \in \real^{n_h}$ to the Transformer encoder module $g$ \citep{2017vaswani+7}, where the hidden representations of agents' observation-action histories are jointly processed. As a result, the $g$ module produces increments, $z^i_t \in \real^{n_h}$. The sum $z^i_t+h^i_t$ is then passed to the second MLP part of the $f^i$ module. We do not use masking or positional encoding (the agents' embeddings are fed to $f^i$ instead). We use a separate Adam optimizer \citep{kingma2017adammethodstochasticoptimization} for the Transformer parameters, while the rest of the architecture is trained with the originally proposed RMSProp optimizer \citep{Tieleman2012}. 
The weights of the output layer of the top Transformer module are initialized with zeros, so initially this module does not affect the agents' estimates of the individual Q-functions, and communication from the self-attention mechanism is gradually incorporated into these estimates.
Consequently, the estimates of the individual Q-functions of each agent are based not only on its own observation-action history, but on the observation-action history of all agents.
The module $g$ learns with the module $f$ via a Q-learning-based algorithm (we use Double DQN \citep{doubleqn}).
\subsection{\ourall{}: tutoring policies with no communication}
\label{subsection:DDCA}

To remain in the CTDE paradigm, we distill in an online manner the policy resulting from the above centralized training into a policy that receives only local information. We refer to the overall architecture as \ourall{} (Decentralized Distillation of Centralized Agents; Fig.~\ref{fig:architecture}). We use a standard MLP-GRU-MLP architecture \citep{2022li+7} to represent the distilled policy $\pi^i_t$, which depends only on the agent's observation-action history ($\tau^i_t$).
We train the distilled policy using the individual Q-function of the agent ($\est Q^i_t$), which, however, depends on the observation–action histories of all agents ($\tau_t \in (\Omega ^\agents \times \actions)^*$). We derive the centralized policy from the individual estimates via argmax, then train the distilled policy using cross-entropy loss. Thus, the loss for the $i$-th decentralized agent in time step $t$ takes the form
\[L^i_t(\theta)\ = D_{KL} \left( e_a || \pi^i_\theta(\cdot | \tau^i_t) \right),\]
where $a = \arg\max \left(\est Q^i_t(\tau_t, \cdot) \right) \in \actions^i$ (the argmax breaks ties and the gradient does not flow through it), $e_a \in \mathbb{P}({\actions^i})$ is the probability distribution that chooses $a$ with probability one, $\theta$ is the parameter vector of the policy, and $D_{KL}$ denotes the Kullback-Leibler divergence. The loss is equivalent to the cross-entropy (negative log-likelihood) loss.
In practice, we share the parameters of $\pi^i$ among the agents, just as is done for the $\est Q^i$ \citep{2022li+7} (centralized and decentralized architectures have separate agents).
To counteract the imitation gap, we gradually introduce samples from the decentralized policy into the replay buffer. In time step $t$ of the training, we decide whether to collect actions using the centralized policy (with probability $p_t$) or using the decentralized policy (with probability $1-p_t$). Then, we act $\epsilon$-greedily according to the chosen policy. It is worth noting that the decision is made for the entire group of agents rather than independently for each agent. 
We find that exponentially annealing $p_t$ yields good performance. We choose
\[p_t = \max(\epsilon_s, \exp({-t}/{scale})),\]
where $scale = \frac{-time}{\ln(\epsilon_s)}$, $\epsilon_s$ is the final value and $time$ is the annealing period. We set $time$ to roughly $\frac{3}{4}$ of the training length and $\epsilon_s$ to $1\%$ for StarCraft-based environments. For Hallway, we observe that choosing $time$ to $60 \%$ at the training time, and $\epsilon_s = 5\%$ allows us to achieve a score of almost $100\%$. Different settings also yield satisfactory results (the score is in the range $60\%-90\%$), and we present hyperparameter sensitivity in the Appendix.
To adjust the centralized policy to the target conditions, we change the Double DQN learning procedure. In Double DQN, the action for the target network is chosen via argmax over the main Q-network. We choose with probability $0.5$ the argmax over the centralized Q-network and with probability $0.5$ the argmax over the distilled policy. This makes the agents aware that in the future the action might be selected by the decentralized policy instead of the better-informed centralized one, thus reducing the imitation gap. We call this mutual adjustment of policies {\it tutoring}. 
We present pseudocode of the whole procedure in the Appendix.

\section{Experiments}

In this section, we present the experimental setup, results, and ablations.

\subsection{Experimental setting}
We use QMIX \citep{2018rashid+5} as the mixing network.
We adopt the basic settings from NA$^2$Q \citep{2023liu+2} and extend their codebase by integrating the original implementations of MAIC, MASIA, and CADP and by implementing \ourcom{} and \ourall{}. The algorithms are trained using the Double DQN \citep{doubleqn} algorithm, with \ourall{} target correction using the adjustment described in the Method section.
Each experiment is run for a fixed number of training episode steps -- 2.0 million for Level-Based Foraging \citep{papoudakis2021foraging} (results presented in the Appendix) and Hallway, 2.5 million for simpler SMAC and SMACv2 environments, and 7.5 million for the hardest SMAC and SMACv2 environments, interleaved with 32 test episodes every 10000 steps. Each experiment is repeated for 5 seeds. We report the average percentage of wins in test games, along with the standard deviations on the plots. Curves are smoothed as in the NA$^2$Q repository \citep{2023liu+2}. We provide interquantile mean plots, a final win rate table, and details of hyperparameter settings and experimental setup in the Appendix.

\akapit{\ourcom{} hyperparameters} We observe that \ourcom{} is sensitive to the Transformer hyperparameters, depending on the scenario's difficulty. The hyperparameters are the number of stacked Transformer encoders and the dimension of their positional feedforward network block. We divide the scenarios into three groups: easy, medium, and hard scenarios. The grouping and the corresponding values are presented in the Appendix. For the hardest SMAC and all SMACv2 maps, we use a consistent architecture of 3 Transformer layers and 512 dimension of the feedforward block.
We use $d_{model} = 64$ in the Transformer architecture (equal to the RNN size).

\paragraph{StarCraft Multi-Agent Challenge.}

\begin{figure}[t!]
  \centering
  \includegraphics[width=0.49\textwidth]{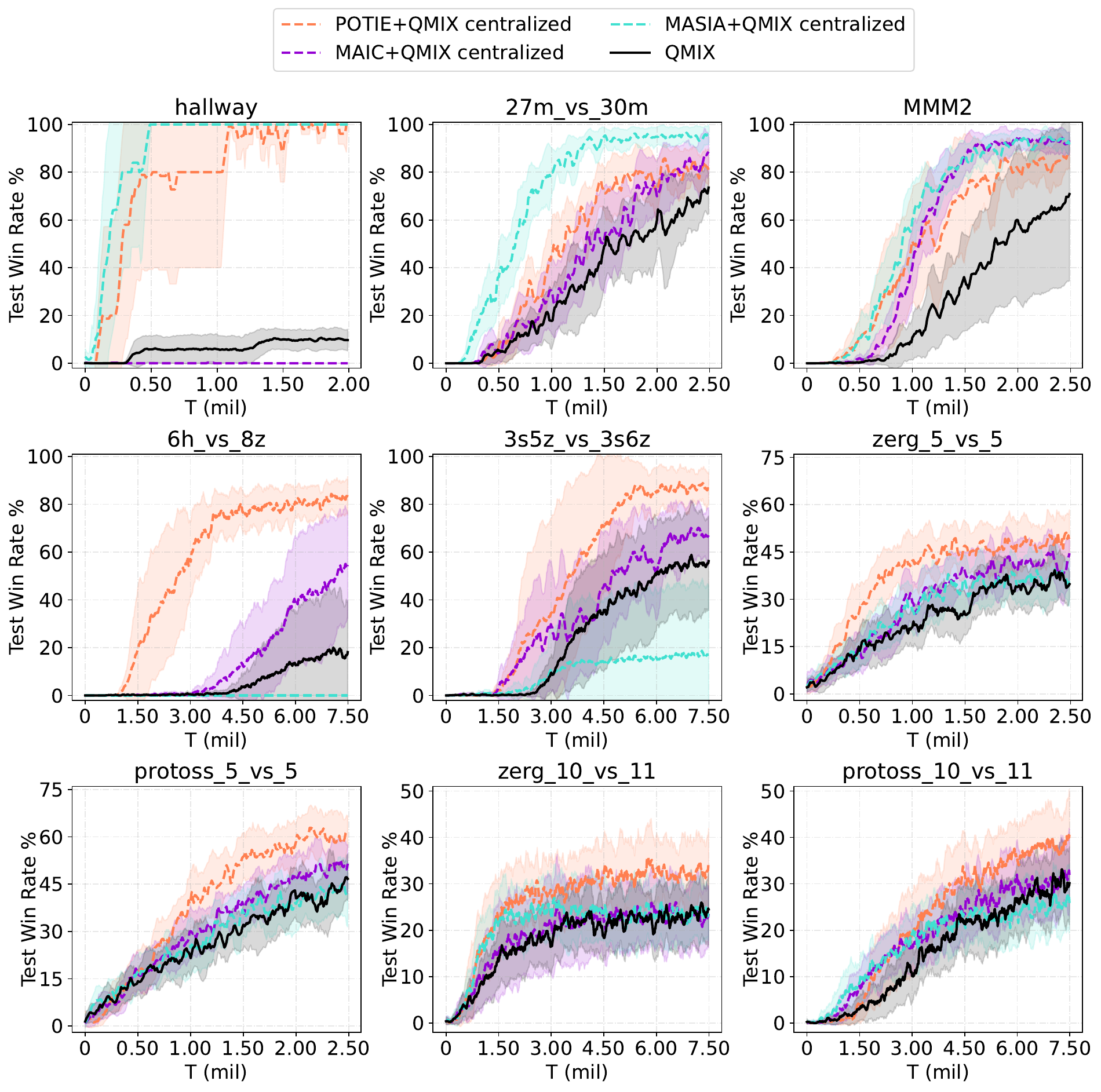}
  \caption{Comparison between \ourcom{}, MAIC, and MASIA on Hallway, SMAC, and SMACv2. We observe that \ourcom{}, as a communication method, outperforms the other methods in most cases, especially in the hard scenarios.}
  \label{fig:MACTAS_comm}
\end{figure}

\begin{figure}[t!]
  \centering
  \includegraphics[width=0.36\textwidth]{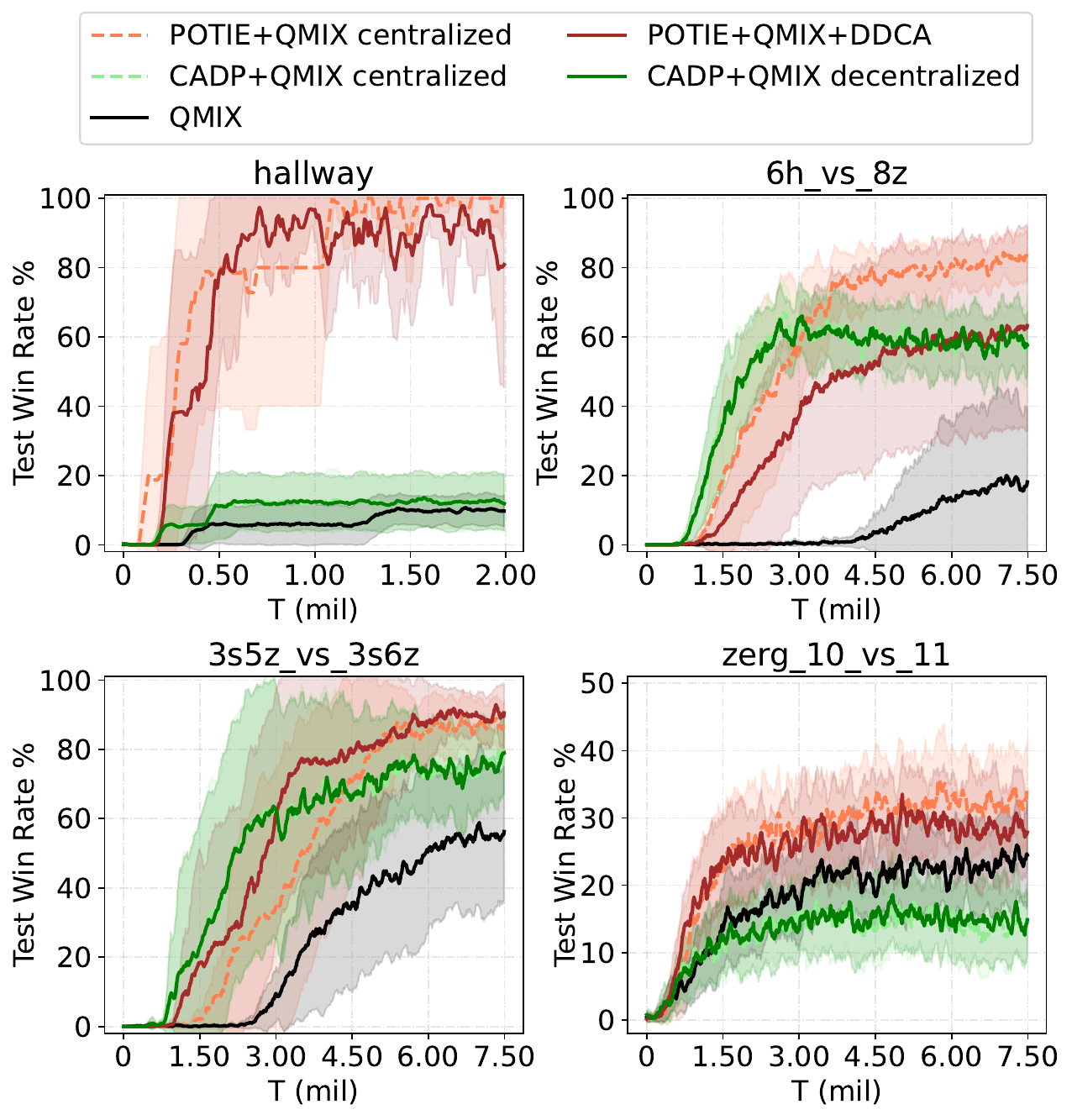}
  \caption{Comparison between \ourall{} using \ourcom{} and CADP on the hardest scenarios. \ourall{} eventually matches the performance of \ourcom{} in most cases (apart from the 6h\_vs\_8z map, where it achieves final performance similar to that of CADP). In particular, we solve Hallway under the CTDE paradigm.}
  \label{fig:MACTAS_vs_CADP}
\end{figure}

\begin{figure}[t!]
  \centering
  \includegraphics[width=0.41\textwidth]{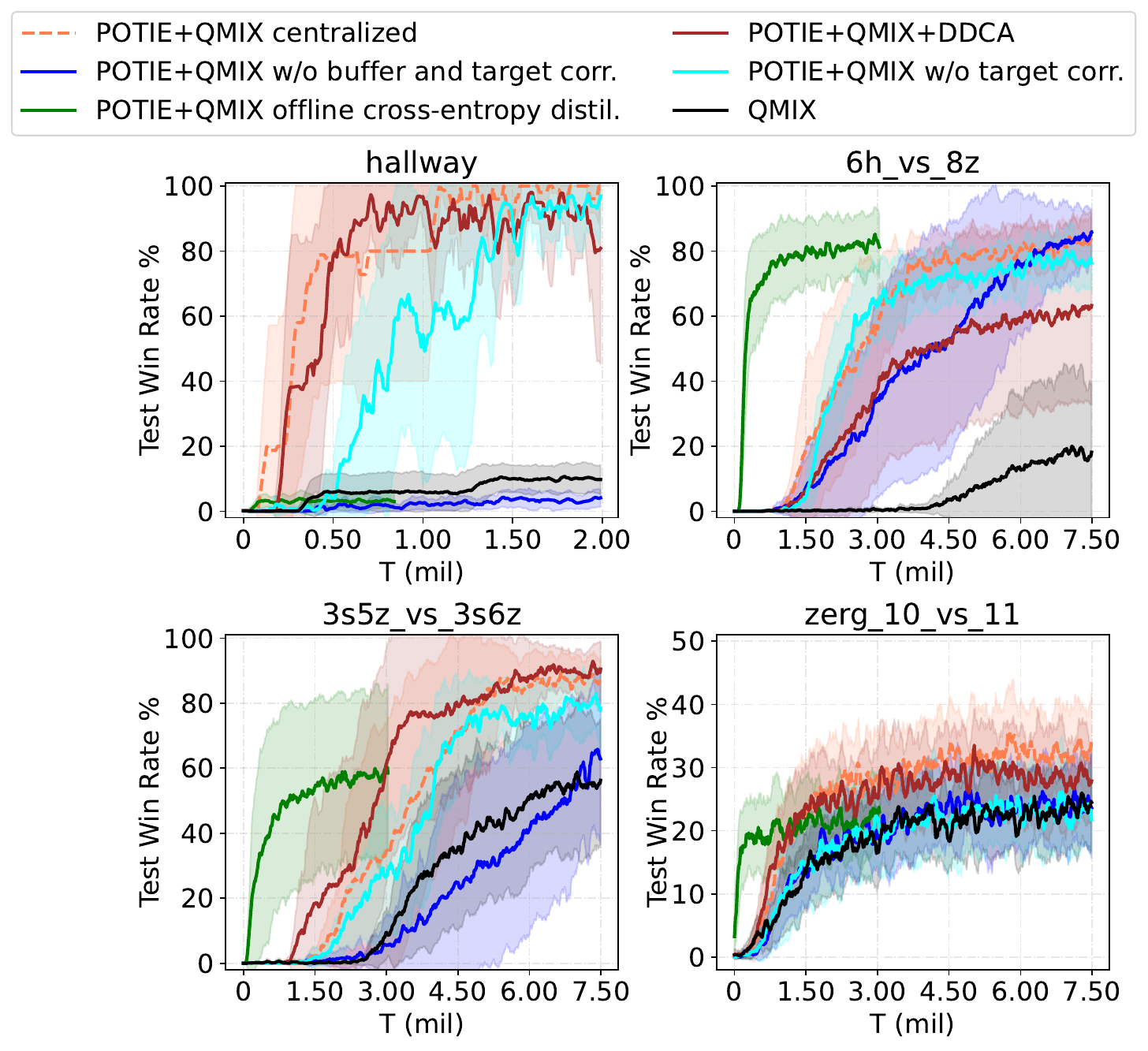}
  \caption{Analysis of the imitation gap corrections and offline cross-entropy distillation. We observe that these corrections improve results on most of the challenging maps (the exceptions are the combined target and replay buffer corrections for the 6h\_vs\_8z map and the replay buffer correction for the zerg\_10\_vs\_11 map). In particular, these corrections allow us to solve the Hallway task.}
  \label{fig:distirbution_shift_correction}
\end{figure}

\begin{figure}[t!]
  \centering
  \includegraphics[width=0.36\textwidth]{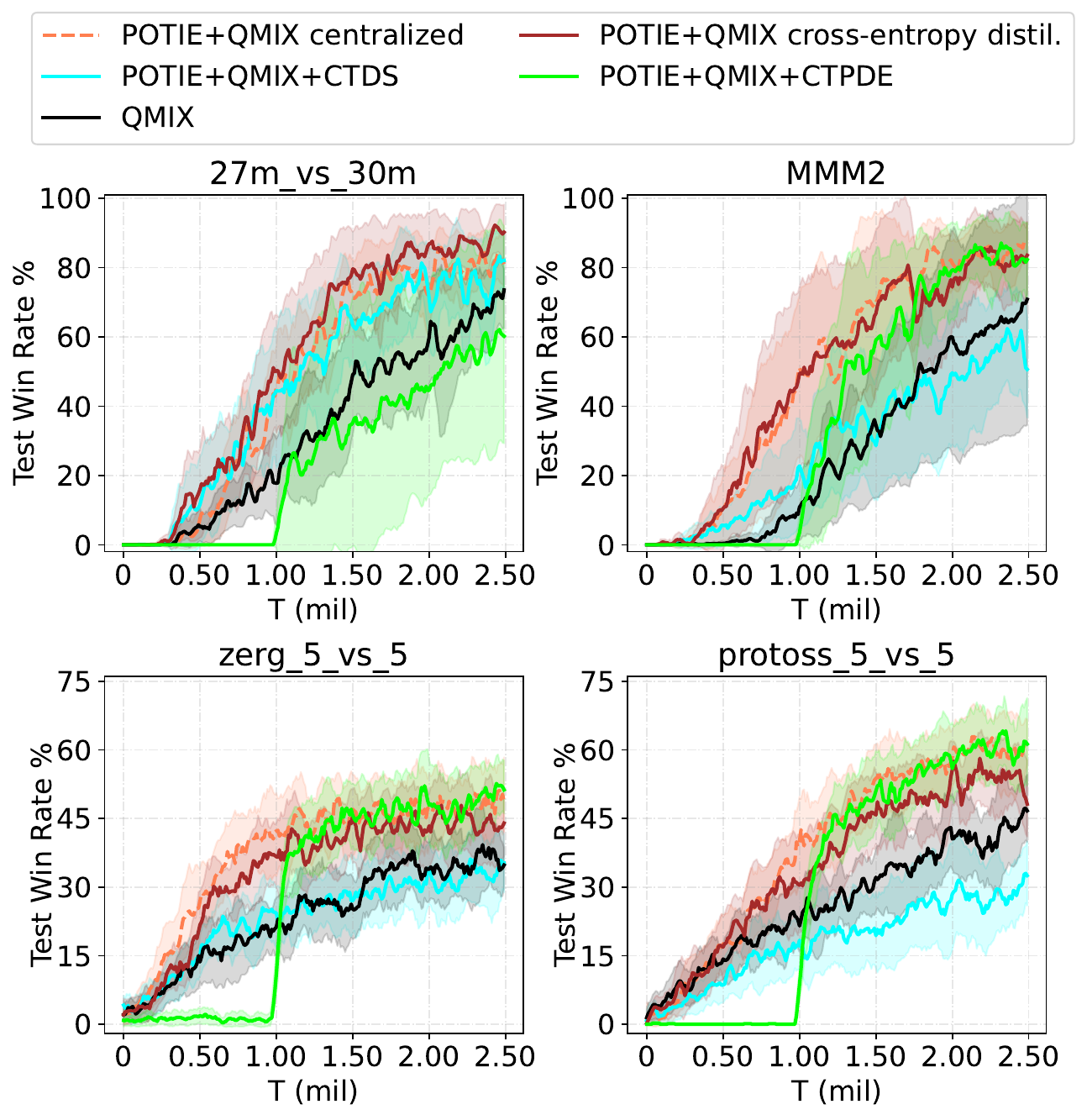}
  \caption{Comparison between cross-entropy-based distillation, CTPDE-based (cross-entropy and KL on the soft policy derived from the Q-function with a one-million-step warm-up), and CTDS-based (MSE) distillation with \ourcom{} as the backbone. The cross-entropy approach visibly outperforms the MSE approach. CTPDE achieves good final performance on all maps except 27m\_vs\_30m.}
  \label{fig:MACTAS_MSE}
\end{figure}

The StarCraft Multi-Agent Challenge (SMAC and SMACv2) is a micromanagement benchmark in which two opposing teams face off across different scenarios (maps), with one team controlled by the RL agent and the other by a built-in algorithm. Maps differ by difficulty, mainly due to an imbalance in the number of controlled units. For our experiments, we selected scenarios from SMAC and SMACv2, which are widely benchmarked in the MARL community and vary in difficulty.
On SMAC \citep{samvelyan19smac}, every map provides the opportunity to find a unique winning strategy. The agents must learn basic skills, such as navigating the battlefield, managing health and shields, or focusing fire. All considered scenarios are imbalanced in favor of the enemies; therefore, simple strategies are insufficient to win the episode.
However, the standard version of SMAC is easily solved by quite simple algorithms. Thus, we adapt the setup from MAIC \citep{2022yuan+5}, in which the episode limits for maps have been shortened, making the benchmark challenging for newer algorithms.
SMACv2 \citep{ellis2023smacv2} is a newer iteration of SMAC, which introduces new maps focused on scenario randomization. Each of the new maps defines the race (the same for both teams) and the number of units in each team, while the exact types and starting positions are randomized according to the configuration. We use the default configuration for each scenario recommended in the SMACv2 repository. 

\paragraph{Hallway.}

Hallway \citep{2019wang+3} is a sparse-reward coordination task with $k$ agents. Agent $i$ is assigned a corridor of length $m_i$, is initialized in this corridor, and can move either right (except at the corridor's right end) or left. There is a special common state $s_g$ to which each agent can move by taking a step to the left from the left end of its corridor. The agents' goal is to move simultaneously to the state $s_g$. We use $k = 3, m_1 = 2, m_2 = 6, m_3 = 10$.

\subsection{Results}
Among architectures with communication, Fig.~\ref{fig:MACTAS_comm} shows \ourcom{} performing very well in the hard scenarios. It also solves Hallway, where MAIC fails. Moreover, as Fig.~\ref{fig:MACTAS_vs_CADP} demonstrates, \ourall{} with \ourcom{} as the centralized tutor often performs similarly to the tutor alone on the particularly hard scenarios, including Hallway.

\akapit{Wall-clock comparison} We compare the running time of the algorithms in Table \ref{tab:speed_test}. \ourcom{} has a reasonable runtime compared with other methods.

\subsection{Ablations}

\begin{figure}[t!]
  \centering
  \includegraphics[width=0.36\textwidth]{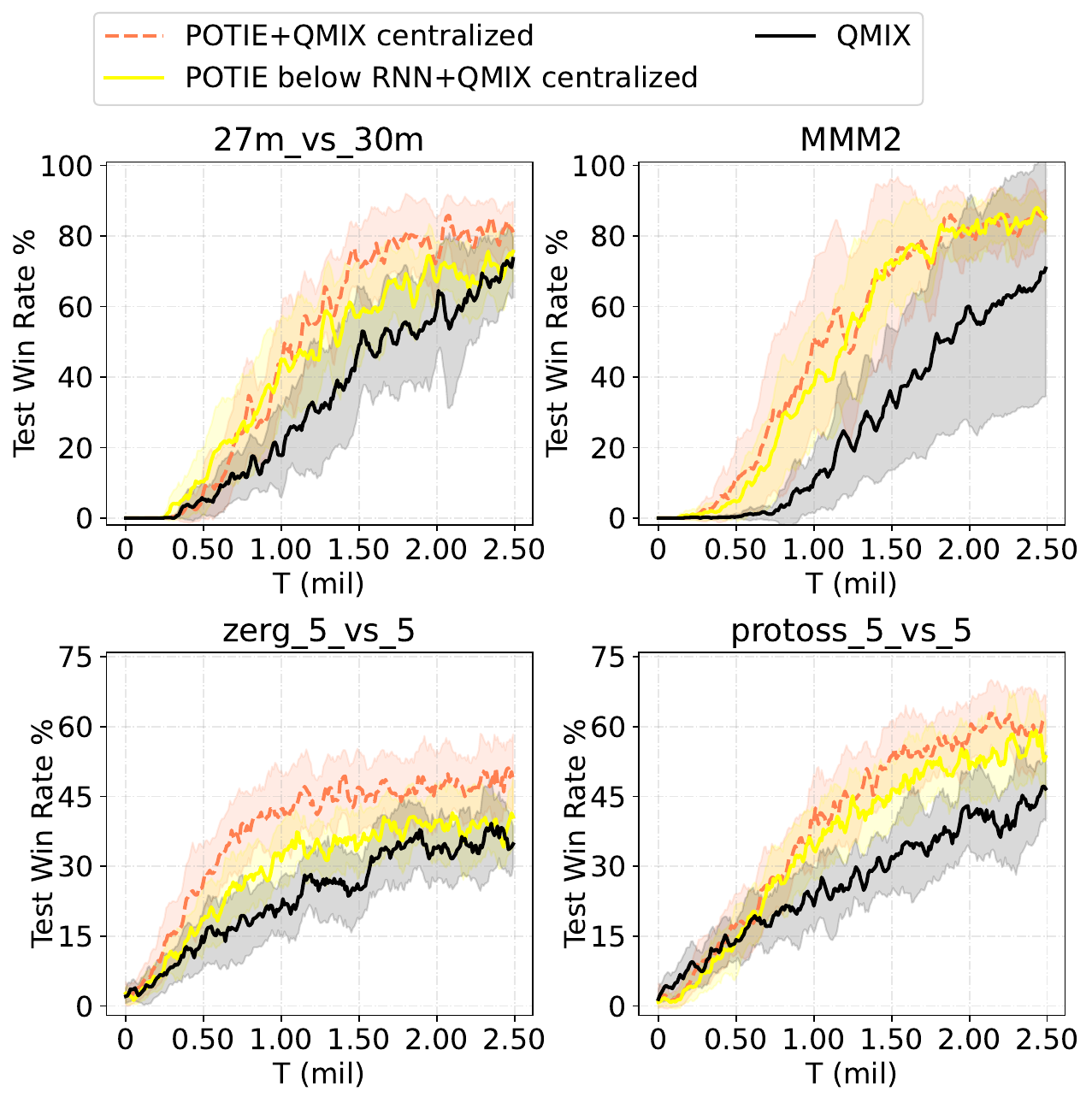}
  \caption{Comparison between \ourcom{} and the same architecture but with the Transformer applied below the recurrent network, thus communicating only current local observations. We observe that using historical information improves the performance.}
  \label{fig:MACTAS_below_RNN}
\end{figure}

\begin{figure}[t!]
  \centering
  \includegraphics[width=0.36\textwidth]{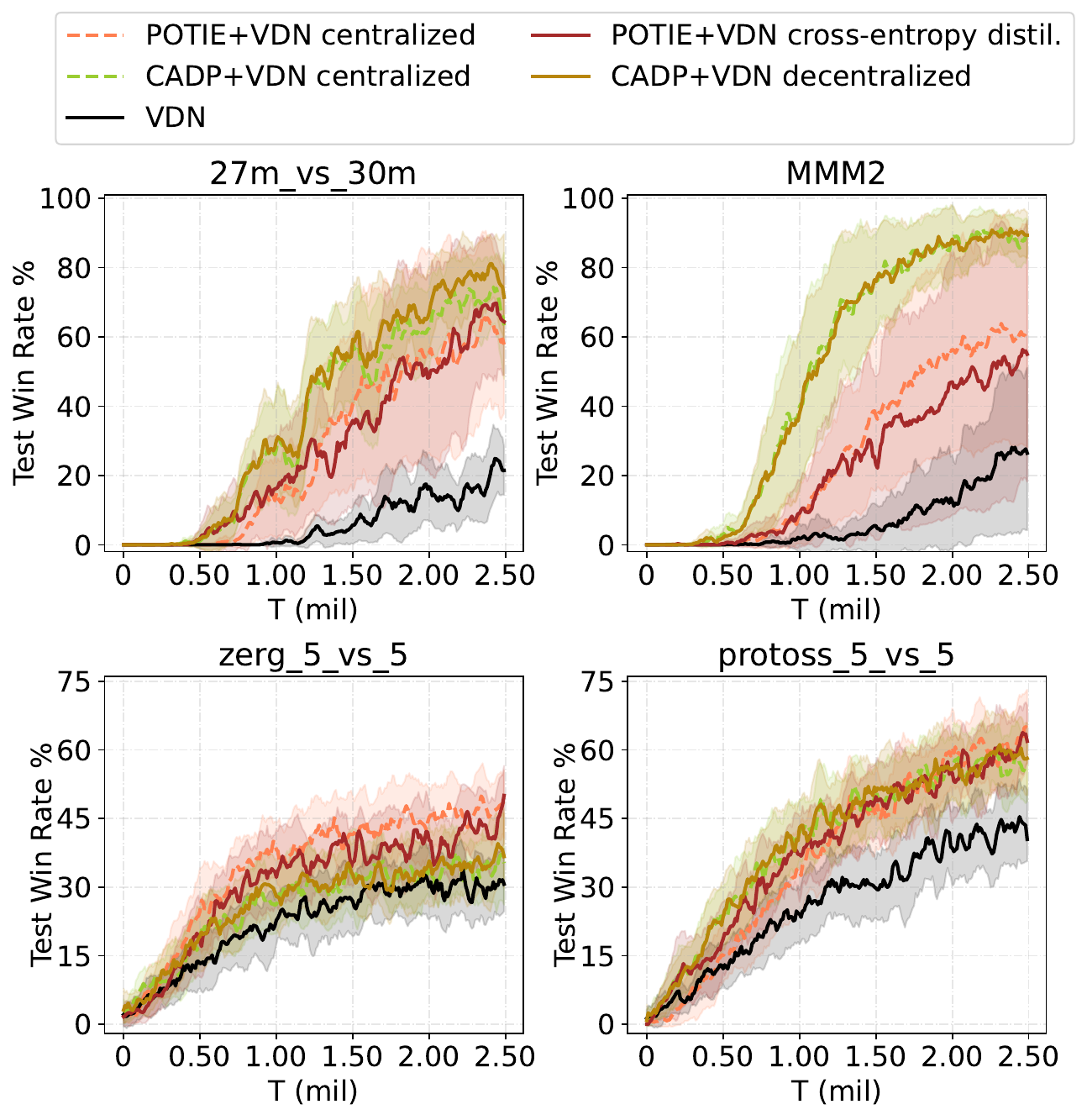}
  \caption{Comparison between \ourcom{} and cross-entropy distillation, and CADP with VDN as the mixer on simple maps. The cross-entropy distillation matches the performance of the centralized counterpart. CADP performs better than the distilled \ourcom{} on SMAC maps and similarly on SMACv2 maps.}
  \label{fig:MACTAS_vs_CADP_VDN}
\end{figure}

\begin{figure}[t!]
  \centering
  \includegraphics[width=0.36\textwidth]{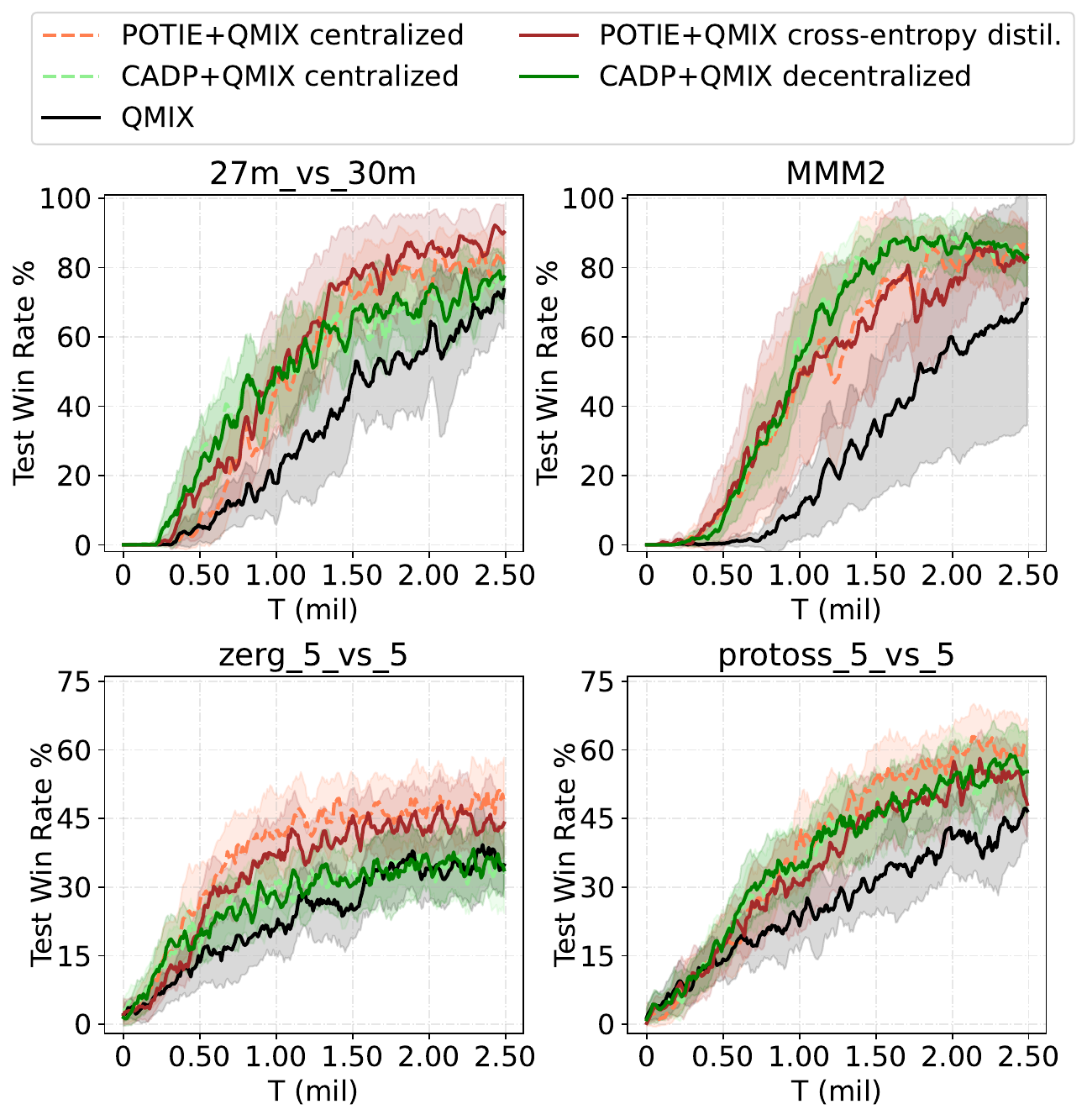}
  \caption{Comparison between \ourcom{} using cross-entropy and CADP with QMIX as the mixer on simple maps. The cross-entropy distillation matches the performance of the centralized counterpart. CADP performs similarly to the distilled \ourcom{}, apart from the zerg\_5\_vs\_5 map, where the cross-entropy distillation performs better.}
  \label{fig:MACTAS_vs_CADP_QMIX_simple}
\end{figure}

\begin{figure}[t!]
  \centering
  \includegraphics[width=0.36\textwidth]{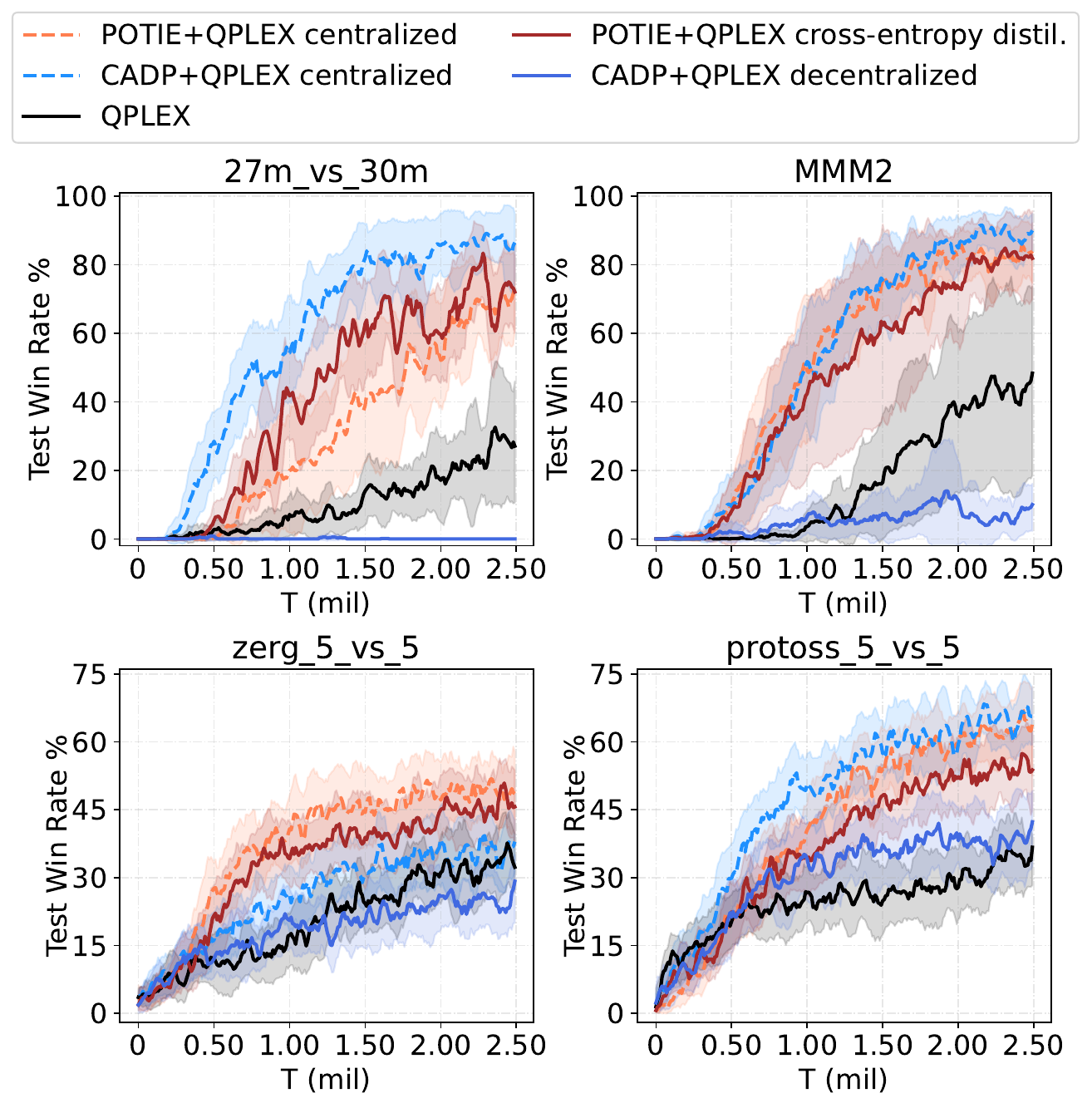}
  \caption{Comparison between \ourcom{} using cross-entropy and CADP with QPLEX as the mixer on simple maps. We observe that the CADP method fails with this mixer and usually does not outperform the bare QPLEX, while both \ourcom{} and cross-entropy distillation prevail.}
  \label{fig:MACTAS_vs_CADP_QPLEX}
\end{figure}

\begin{figure}[t!]
  \centering
  \includegraphics[width=0.36\textwidth]{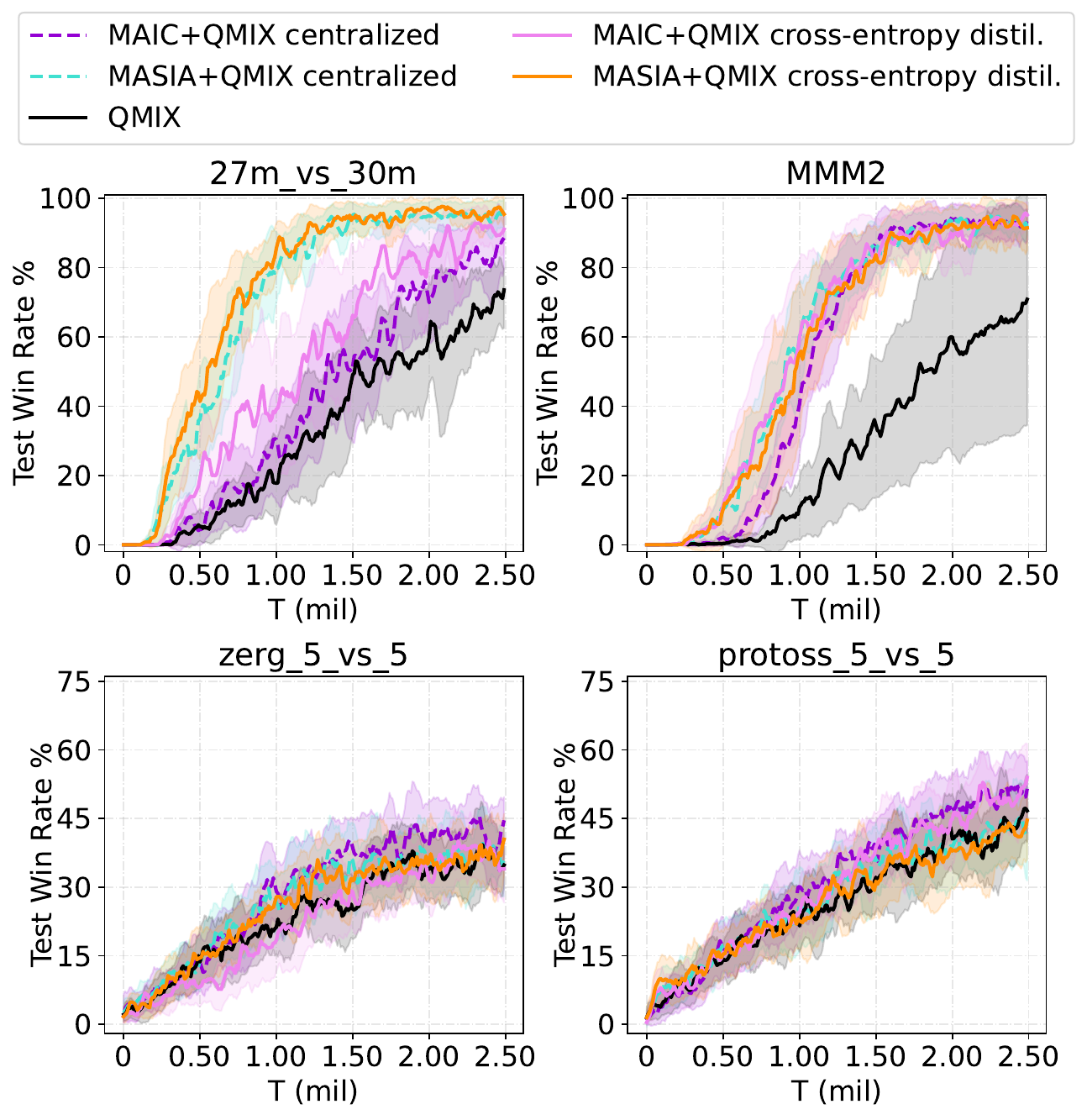}
  \caption{The cross-entropy distillation applied to MAIC and MASIA on simple maps. We observe close consistency between the centralized and distilled versions (the only exception is MAIC on the zerg\_5\_vs\_5 map).}
  \label{fig:MAIC_MASIA_distillation}
\end{figure}

\akapit{Using historical information in \ourcom{}} We verify the gain that comes from communicating historical information as well instead of communicating only current local observations. The results are presented in Fig. \ref{fig:MACTAS_below_RNN} and show the advantage of communicating historical information.

\akapit{Different distillation methods}
The \ourall{} distillation method is based on the cross-entropy loss. We compare this approach with the CTPDE-like distillation and CTDS-like (MSE) distillation with \ourcom{} as the backbone architecture.
The results are presented in Fig.~\ref{fig:MACTAS_MSE}. We observe that the MSE-based approach underperforms the KL-divergence-based approaches, while the CTPDE-like approach yields comparable results apart from one map.

\akapit{Imitation gap corrections and offline distillation}
We study the impact of the imitation gap corrections in the hardest scenarios.
Although in this paper we consider online distillation, we also conduct an experiment using offline distillation. We train \ourcom{}, then distill it for $40\%$ of the main run's length, enough to reach a plateau. The combined results are presented in Fig.~\ref{fig:distirbution_shift_correction}. We observe that \ourall{} often outperforms ablated versions; an interesting case is the \verb|6h_vs_8z| map, where the target correction seems to hinder the learning process.

\akapit{Cross-entropy distillation on simple maps}
We found that on simple maps, cross-entropy distillation alone is usually enough to match the performance of the centralized algorithms. We support this claim with a study of distillation from \ourcom{} with QMIX \citep{2018rashid+5}, VDN \citep{2018sunehag+10}, and QPLEX \citep{2021wang+4} and with distillation from MAIC \citep{2022yuan+5} and MASIA \citep{Guan2022masia}. Additionally, we compare the performance of CADP \citep{2025zhou+6CADP} and \ourcom{} with the aforementioned mixers on these simple maps. The results are presented in Fig.~\ref{fig:MACTAS_vs_CADP_VDN}-\ref{fig:MAIC_MASIA_distillation}.

\section{Conclusions}

We introduced \ourcom{}, a Transformer-based information-sharing method for MARL, and \ourall{}, a~MARL tutoring method that operates within the CTDE paradigm. We showed that \ourcom{} usually outperforms strong communication-based baselines, namely MAIC and MASIA, across a diverse range of scenarios. Moreover, \ourall{} tutoring often matches its centralized counterpart while outperforming CADP, a strong CTDE baseline based on a Transformer and communication pruning. In particular, we solve the Hallway task without test-time communication, which, to the best of our knowledge, has not been done before.

{\bf Limitations:} 
\ourcom{} needs a smaller architecture on simple tasks and a larger Transformer on hard ones; we choose the number of Transformer layers and the Transformer feedforward dimension differently for easy, medium, and hard scenarios.
Even though the parameter count of \ourcom{} is constant and the wall-clock measurements are comparable to the baseline methods, the attention mechanism has quadratic computational complexity. This may necessitate the use of linear attention \citep{pmlr-v119-katharopoulos20a, wang2020linformerselfattentionlinearcomplexity} in future work to apply \ourcom{} to large-scale systems.
We observe that on the \verb|6h_vs_8z| map \ourall{} performs worse than the version without imitation gap corrections and comparably to CADP. However, in the remaining hard tasks, including Hallway, these corrections provide substantial gains.

\begin{table}[ht]
\centering
\small
\setlength{\tabcolsep}{4pt}
\begin{tabular}{llrr}
\toprule
\textbf{Map} & \textbf{Algorithm} & \textbf{Average [h]}& \textbf{Std [h]} \\
\midrule
\multirow{5}{*}{zerg\_10\_vs\_11}
  & MAIC+QMIX                   & 18.07 & 0.17 \\
  & MASIA+QMIX                  & 10.41 & 0.07 \\
  & CADP+QMIX                   & 11.88 & 0.02 \\
  & POTIE+QMIX (ours)           & 12.86 & 0.13 \\
  & QMIX                        &  6.48 & 0.07 \\
\midrule
\multirow{5}{*}{zerg\_5\_vs\_5}
  & MAIC+QMIX                   & 13.82 & 0.06 \\
  & MASIA+QMIX                  &  9.05 & 0.01 \\
  & CADP+QMIX                   &  8.55 & 0.06 \\
  & POTIE+QMIX (ours)           & 12.92 & 0.16 \\
  & QMIX                        &  5.44 & 0.29 \\

\bottomrule
\end{tabular}
\caption{
Runtime results for each map and algorithm over one million environment steps. POTIE is based on 3 Transformer layers with a feedforward dimension of 512.
}
\label{tab:speed_test}
\end{table}

\section*{Acknowledgements}

We gratefully acknowledge Polish high-performance computing infrastructure PLGrid (HPC Center: ACK Cyfronet AGH) for providing computer facilities and support within computational grant no. PLG/2025/018560.
We gratefully acknowledge Polish high-performance computing infrastructure PLGrid (HPC Center: ACK Cyfronet AGH) for providing computer facilities and support within computational grant no. PLG/2026/019802.
We gratefully acknowledge Polish high-performance computing infrastructure PLGrid (HPC Center: WCSS) for providing computer facilities and support within computational grant no. PLG/2026/019115.
We thank the University of Warsaw for providing access to the computing infrastructure.

\bibliography{aaai2027}

\end{document}


\maketitle
\appendix
\input{appendix}
\bibliography{aaai2027}

%% file: defs.tex
\usepackage{calc}

\usepackage{amsmath,amsfonts}

\def\agents{\mathbb{N}}
\def\states{\mathbb{S}}
\def\actions{\mathbb{A}}

\def\P{\mathcal{P}}
\def\R{\mathcal{R}}
\def\O{\mathcal{O}}
\def\T{\mathcal{T}}
\def\V{\mathcal{V}}

\newcommand\est{\widehat}

\newcommand\real{\mathbb{R}}

\newcommand\akapit[1]{\paragraph{#1.}}

\def\ourcom{POTIE}
\def\ourall{DDCA}

%% file: appendix.tex
\section{Extended Related Work}
\label{sec:rel-work:comm}

\subsection{Value decomposition methods}

There are many approaches to enhance the mixing architectures to solve the value decomposition problem.
Weighted QMIX \citep{2020rashid+3} refines the QMIX approach \citep{2018rashid+5} by assigning special weights to optimal joint actions during training. QTRAN \citep{2019son+4} introduces an auxiliary module that incorporates the joint action to correct the decomposed value. Attention mechanisms are leveraged in Qatten \citep{2020yang+6} to construct a more flexible mixing network. Additional extensions include CDS \citep{2021li+5}, which adjusts agent exploration through the mixing network; DVD \citep{2022li+7}, which incorporates agents’ internal states into value aggregation; SHAQ \citep{2022wang+3}, which augments the mixing network using principles from Shapley value theory; and NA$^2$Q \citep{2023liu+2}, which adopts a similar duplex dueling strategy to QPLEX \citep{2021wang+4} enhanced with attention mechanisms. 

\subsection{Communication in cooperative MARL}
A fundamental problem in cooperative MARL is coordinating agents' activities. It is addressed with a communication protocol. A number of solutions have been proposed for this purpose: CommNet \citep{2016sukhbaatar+3} introduces differentiable communication by aggregating a learned broadcast vector from each agent, TarMAC \citep{2019das+5} uses signature-based soft attention to direct agent communication towards recipients, NDQ \citep{2019wang+3} applies two forms of regularization -- mutual information maximization and message entropy minimization to optimize message content, GA-Comm \citep{2020liu+5} uses a two-stage attention mechanism to limit the number of messages exchanged between agents, IS \citep{2021kim+2} encodes a trajectory of future actions for each agent and shares it with the other agents, DHCG \citep{2023liu+5} learns a dependency graph between agents and uses it to limit the number of communication edges, TEM \citep{2023guo+2} uses a Transformer to simulate communication resembling email exchange, where agents only directly communicate with teammates nearby, but can forward messages from one to another, ExpoComm \citep{LiExpoComm2025} utilizes exponential graphs to design the communication topology. Also, in MAIC \citep{2022yuan+5}, agents explicitly model their teammates and use these models to bias their value function; MASIA \citep{Guan2022masia} uses an attention-based communication architecture and a decoder module to reconstruct the global state; CommFormer \citep{hu2024commformer} uses an attention-based encoder--decoder architecture and learns a communication graph, and its number of parameters is independent of the number of agents.
In addition, there are solutions in which the agents communicate in natural language, including TWOSOME \citep{2024tan+5}, which uses a large language model (LLM) to evaluate the probabilities of joint action and applies LoRA for efficient PPO training, and Verco \citep{2024li+10}, which separates the output of the LLM into communication and actions, allowing LLM agents to exchange interpretable verbal messages.

\subsection{Attention architectures for MARL}
Attention and Transformer architectures have been applied in MARL algorithms, such as in Qatten \citep{2020yang+6} and NA$^2$Q (QNAM) \citep{2023liu+2} mixing networks, in communication mechanisms such as GA-Comm \citep{2020liu+5}, TEM \citep{2023guo+2}, CommFormer \citep{hu2024commformer}, MASIA \citep{Guan2022masia}, and in CADP \citep{2025zhou+6CADP}, which uses gradually pruned communication in training and turns it off during execution. MAT \citep{wen2022multiagent} treats MARL as a sequence modeling problem and uses a Transformer to solve it. UPDeT \citep{hu2021updet} utilizes a Transformer to produce multi-agent policies compatible with different observation and action configurations. MARIE \citep{zhang2025decentralized} uses a Transformer architecture as a world model for MARL.

\section{Pseudocode}

We show the pseudocode for \ourall{} in Algorithms \ref{alg} and \ref{alg:action_choice}.

\begin{figure}
\begin{minipage}{0.46\textwidth}
    \vspace{-0.25cm}
    \begin{algorithm}[H]
        \small
        \caption{\ourall{} algorithm}
        \label{alg}
        \begin{algorithmic}[1]
            \STATE {\bfseries Input:} Number of training iterations $T$, Episode length $K$, Discount factor $\gamma$, Initial exploration rate $\varepsilon_0$, Target network update frequency $T_{update}$, Initialized main history and Q-networks $\texttt{Hist}, Q$, Initialized target history and Q-networks $\texttt{Hist}_{targ}, Q_{target}$, Initialized main and target mixing networks $\texttt{Mix}, \texttt{Mix}_{targ}$, Initialized decentralized history and policy $\texttt{Hist}_{local}, \pi$ networks.
            \STATE {\bfseries Return:} Trained decentralized policy $\pi$ network.
            \STATE $D \gets \emptyset,\ \varepsilon \gets \varepsilon_0,\ p_{target} \gets p_0 = 0.5,\ g \gets 0$
            \FOR {$t = 1,\ \dots,\ T$}
                \STATE Generate trajectories using the action-selection procedure in Algorithm \ref{alg:action_choice} and update $D, \varepsilon, p$
                \STATE $\big(s_r, (obs_r)_{i=1}^N, (\mathit{action}_r)_{i=1}^N, \mathit{rew}_r, \mathit{done}_r \big)_{r=1}^{K} \sim D$
                \FOR{$r = 1,\ \dots,\ K$}
                    \STATE $h_r \gets \big(s_{1, \dots r}, (obs_{1 \dots r})_{i=1}^N, (\mathit{action}_{1 \dots r})_{i=1}^N\big)$
                    \STATE $(\tau_r)_{i=1}^N \gets \texttt{Hist}(h_r)$,\ \ $(targ\_\tau_r)_{i=1}^N \gets \texttt{Hist}_{targ}(h_r)$
                    \FOR {$i = 1,\ \dots,\ N$}
                        \STATE $(h\_local_r)_i \gets \big((obs_{1 \dots r})_{i}, (\mathit{action}_{1 \dots r})_{i}\big)$
                        \STATE $(\tau\_local_r)_{i} \gets \texttt{Hist}_{local}(h\_local_r)_i$
                    \ENDFOR
                \ENDFOR
                \FOR{$r = 1,\ \dots,\ K-1$}
                    \FOR {$i = 1,\ \dots,\ N$}
                        \STATE $\mathit{chosen\_action}_i \gets (action_r)_{i}$
                        \STATE $\mathit{max\_action}_i \gets \texttt{argmax} Q_i(\tau_{r}, \cdot)$
                        \STATE $rand_i \sim \textit{U}_{[0, 1)}$
                        \IF{$rand_i \leq p_{target}$}
                            \STATE $\mathit{max\_next\_action}_i \gets \texttt{argmax} Q_i(\tau_{r+1}, \cdot)$
                        \ELSE
                            \STATE $\mathit{max\_next\_action}_i \gets \texttt{argmax} \pi_i((\tau\_local_{r+1})_i, \cdot)$
                        \ENDIF
                        \STATE $qvals_i \gets Q((\tau_r)_i, \mathit{chosen\_action}_i)$,\ \ $targ\_qvals_i \gets Q_{target}((targ\_\tau)_{r+1}, \mathit{max\_next\_action}_i)$
                        \STATE Update the policy network via cross-entropy between $\pi_i((\tau\_local_r)_{i}, \cdot)$ and the one-hot encoded detached $\mathit{max\_action}_i$
                    \ENDFOR
                    \STATE $Q_{tot} \gets \texttt{Mix}\big(s_r, (qvals)_{i=1}^N\big)$
                    \STATE $targ\_Q_{tot} \gets \texttt{Mix}_{targ}\big(s_{r+1}, (targ\_qvals)_{i=1}^N\big)$
                    \STATE $target \gets \mathit{rew}_r + \gamma \cdot (1-\mathit{done}_r) \cdot targ\_Q_{tot}$
                    \STATE Update the main network using the MSE between $Q_{tot}$ and the detached target.
                    \STATE $g \gets g + 1$
                    \IF{$g \bmod T_{update} = 0$}
                        \STATE Update the target network
                    \ENDIF
                \ENDFOR 
            \ENDFOR
            
        \end{algorithmic}
    \end{algorithm}
\end{minipage}
\vspace{-0.2cm}
\setcounter{algorithm}{0}
\captionof{algorithm}{The \ourall{} procedure with optional corrections. In practice, most of the \texttt{for} loops are parallelizable.}
\end{figure}

\begin{figure}
\centering
\begin{minipage}{0.46\textwidth}
    \vspace{-0.25cm}
    \begin{algorithm}[H]
        \small
        \caption{The action-selection function used to train \ourall{}}
        \label{alg:action_choice}
        \begin{algorithmic}[1]
        \STATE {\bfseries Input:} Replay buffer $D$, $\varepsilon$, $p$, number of steps $T_{\mathrm{gather}}$, number of agents $N$, distribution of the initial state $\mathcal{P}(s_0)$, transition probability kernel $P$, reward function $R$, observation function $O$, annealing schedule $\texttt{Anneal}\varepsilon(n, \varepsilon)$, exponential annealing schedule $\texttt{AnnealExponentiallyP}(n, p)$.
        \STATE {\bfseries Output:} Updated replay buffer $D$, $\varepsilon$, $p$.
        \STATE $s \sim \mathcal{P}(s_0)$
        \FOR{$n = 1,\ \dots,\ T_{\mathrm{gather}}$}
            \STATE $\varepsilon \gets \texttt{Anneal}\varepsilon\left(n, \varepsilon\right)$
            \STATE $p \gets \texttt{AnnealExponentiallyP}\left(n, p\right)$
            \STATE $\mathit{group\_rand} \sim \mathcal{U}_{[0, 1)}$
            \FOR{$i = 1,\ \dots,\ N$}
                \STATE $\mathit{rand}_i \sim \mathcal{U}_{[0, 1)}$
                \IF{$\mathit{rand}_i < \varepsilon$}
                    \STATE $\mathit{action}_i \gets \texttt{random action}$
                \ELSIF{$\mathit{group\_rand} < p$}
                    \STATE $\mathit{action}_i \gets \operatorname{argmax} Q_i$
                \ELSE
                    \STATE $\mathit{action}_i \gets \operatorname{argmax} \pi_i$
                \ENDIF
            \ENDFOR
            \IF{Episode ends in $s$}
                \STATE $\mathit{done} \gets TRUE$
                \STATE $s^\prime \sim \mathcal{P}(s_0)$
                \STATE $\mathit{rew} \gets R\big(s, (\mathit{action}_i)_{i=1}^N, END\big)$
            \ELSE
                \STATE $\mathit{done} \gets FALSE$
                \STATE $s^\prime \sim P\big(\cdot \mid s, (\mathit{action}_i)_{i=1}^N\big)$
                \STATE $\mathit{rew} \gets R\big(s, (\mathit{action}_i)_{i=1}^N, s^\prime\big)$
            \ENDIF
            \FOR{$i = 1,\ \dots,\ N$}
                \STATE $\mathit{obs}_i \gets O(i, s)$
            \ENDFOR
            \STATE $D \gets \texttt{UpdateBuffer}$ $\big(D, s, (\mathit{action}_i)_{i=1}^N, s^\prime, (\mathit{obs}_i)_{i=1}^N, \mathit{rew}, \mathit{done}\big)$
            \STATE $s \gets s^\prime$
        \ENDFOR
        \end{algorithmic}
    \end{algorithm}
\end{minipage}
\vspace{-0.2cm}
\end{figure}

\section{Extended experimental results}
\label{sec:extended_results}

We present additional experimental results in this section.

\subsection{Tabular results}

\begin{table*}[t]
\centering
\begin{tabular}{lcccc}
\toprule
Map & MAIC & MASIA & QMIX & POTIE (Ours) \\
\midrule
27m\_vs\_30m & 87.1 $\pm$ 9.8 & \textbf{95.4 $\pm$ 2.5} & 71.5 $\pm$ 7.0 & 82.4 $\pm$ 4.6 \\
3s5z\_vs\_3s6z & 66.4 $\pm$ 10.4 & 17.1 $\pm$ 32.9 & 56.2 $\pm$ 20.3 & \textbf{85.6 $\pm$ 5.4} \\
6h\_vs\_8z & 54.6 $\pm$ 24.4 & 0.0 $\pm$ 0.0 & 18.1 $\pm$ 26.8 & \textbf{84.8 $\pm$ 3.2} \\
MMM2 & 91.9 $\pm$ 3.8 & \textbf{92.1 $\pm$ 3.5} & 69.6 $\pm$ 39.1 & 87.4 $\pm$ 3.0 \\
hallway-3a-1g & 0.0 $\pm$ 0.0 & \textbf{100.0 $\pm$ 0.0} & 9.7 $\pm$ 4.6 & 96.1 $\pm$ 8.6 \\
protoss\_10\_vs\_11 & 31.9 $\pm$ 3.6 & 25.9 $\pm$ 3.8 & 30.1 $\pm$ 6.4 & \textbf{40.6 $\pm$ 3.1} \\
protoss\_5\_vs\_5 & 49.8 $\pm$ 3.7 & 42.1 $\pm$ 7.6 & 47.1 $\pm$ 3.9 & \textbf{61.1 $\pm$ 3.3} \\
zerg\_10\_vs\_11 & 23.5 $\pm$ 4.0 & 22.9 $\pm$ 2.9 & 24.5 $\pm$ 4.9 & \textbf{34.8 $\pm$ 4.8} \\
zerg\_5\_vs\_5 & 42.0 $\pm$ 3.4 & 35.0 $\pm$ 4.0 & 33.9 $\pm$ 2.3 & \textbf{49.1 $\pm$ 2.0} \\
\bottomrule
\end{tabular}
\caption{Test win rate (\%), mean $\pm$ std. Best mean per map in bold. We observe that \ourcom{} outperforms baselines in most cases. In three cases, MASIA is the best algorithm, but it fails on some maps (6h\_vs\_8z, 3s5z\_vs\_3s6z). MAIC is a more robust baseline on the StarCraft-based environments, but fails on Hallway.}
\label{tab:final_results_centralized}

\vspace{1em}

\begin{tabular}{lccc}
\toprule
Map & QMIX & CADP & DDCA (Ours) \\
\midrule
3s5z\_vs\_3s6z & 56.2 $\pm$ 20.3 & 79.0 $\pm$ 11.3 & \textbf{90.5 $\pm$ 9.2} \\
6h\_vs\_8z & 18.1 $\pm$ 26.8 & 57.6 $\pm$ 7.5 & \textbf{63.2 $\pm$ 32.4} \\
hallway-3a-1g & 9.7 $\pm$ 4.6 & 12.4 $\pm$ 9.0 & \textbf{80.4 $\pm$ 22.5} \\
zerg\_10\_vs\_11 & 24.5 $\pm$ 4.9 & 14.9 $\pm$ 3.3 & \textbf{27.9 $\pm$ 3.7} \\
\bottomrule
\end{tabular}
\caption{Test win rate (\%), mean $\pm$ std. Best per map in bold. We observe that \ourall{} outperforms the baselines in 3 cases and yields results similar to CADP on the 6h\_vs\_8z map.}
\label{tab:final_results_decentralized}
\end{table*}

We present the tabular results for the comparison of the performance of \ourcom{}
Table \ref{tab:final_results_centralized} and of \ourall{} in Table \ref{tab:final_results_decentralized}. We compute the mean value across the last five evaluation steps, and compute average and standard deviations across the seeds. Our methods yield the best results in most cases.

\subsection{\ourcom{} performance against baselines}

\begin{figure*}
  \centering
  \includegraphics[width=0.79\textwidth]{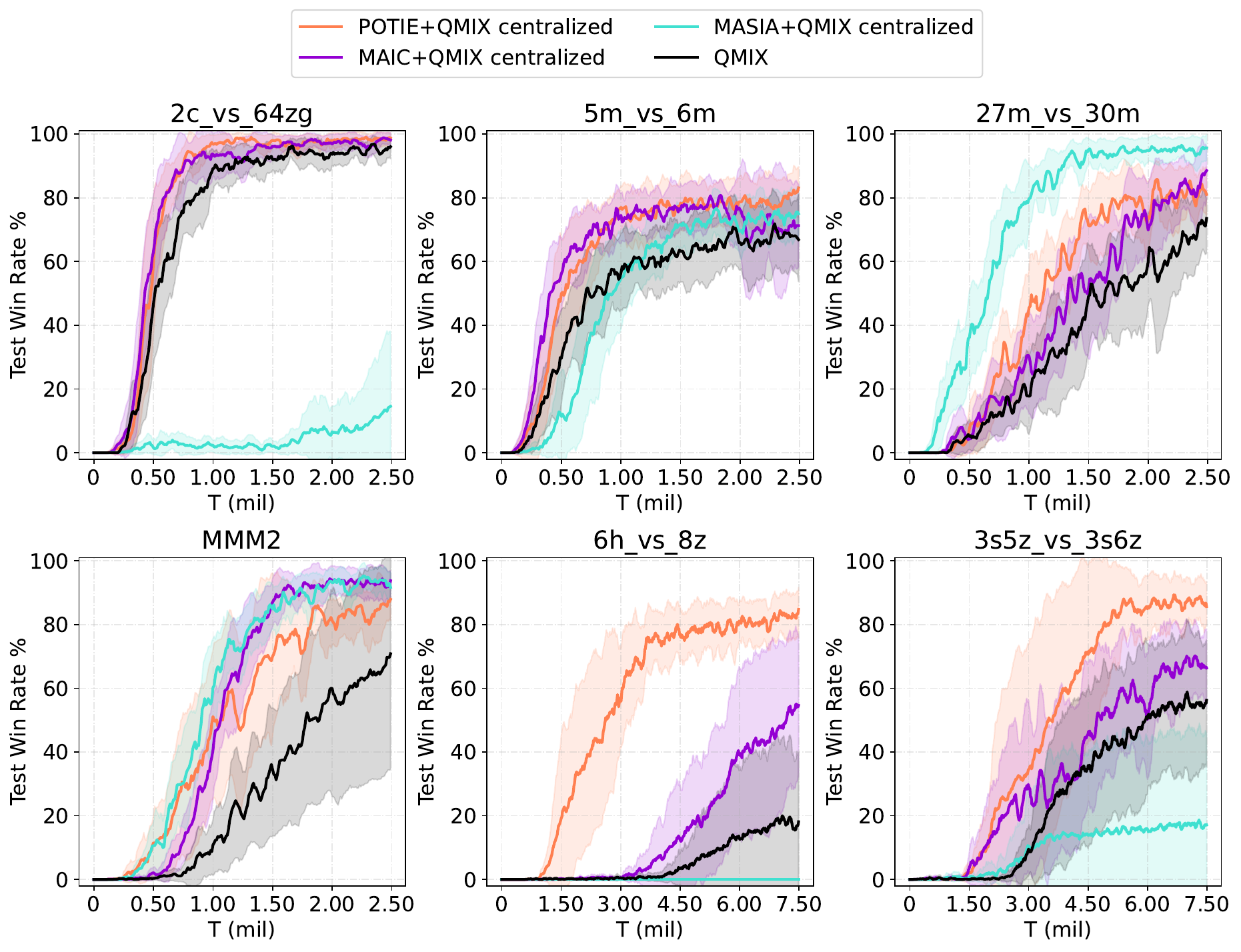}
  \caption{Comparison of different communication algorithms with the QMIX mixer on a variety of SMAC maps. We observe particularly good performance of \ourcom{} on the two most challenging maps.}
  \label{fig:POTIE_vs_comm_qmix_smacv1}
\end{figure*}

\begin{figure*}
  \centering
  \includegraphics[width=0.79\textwidth]{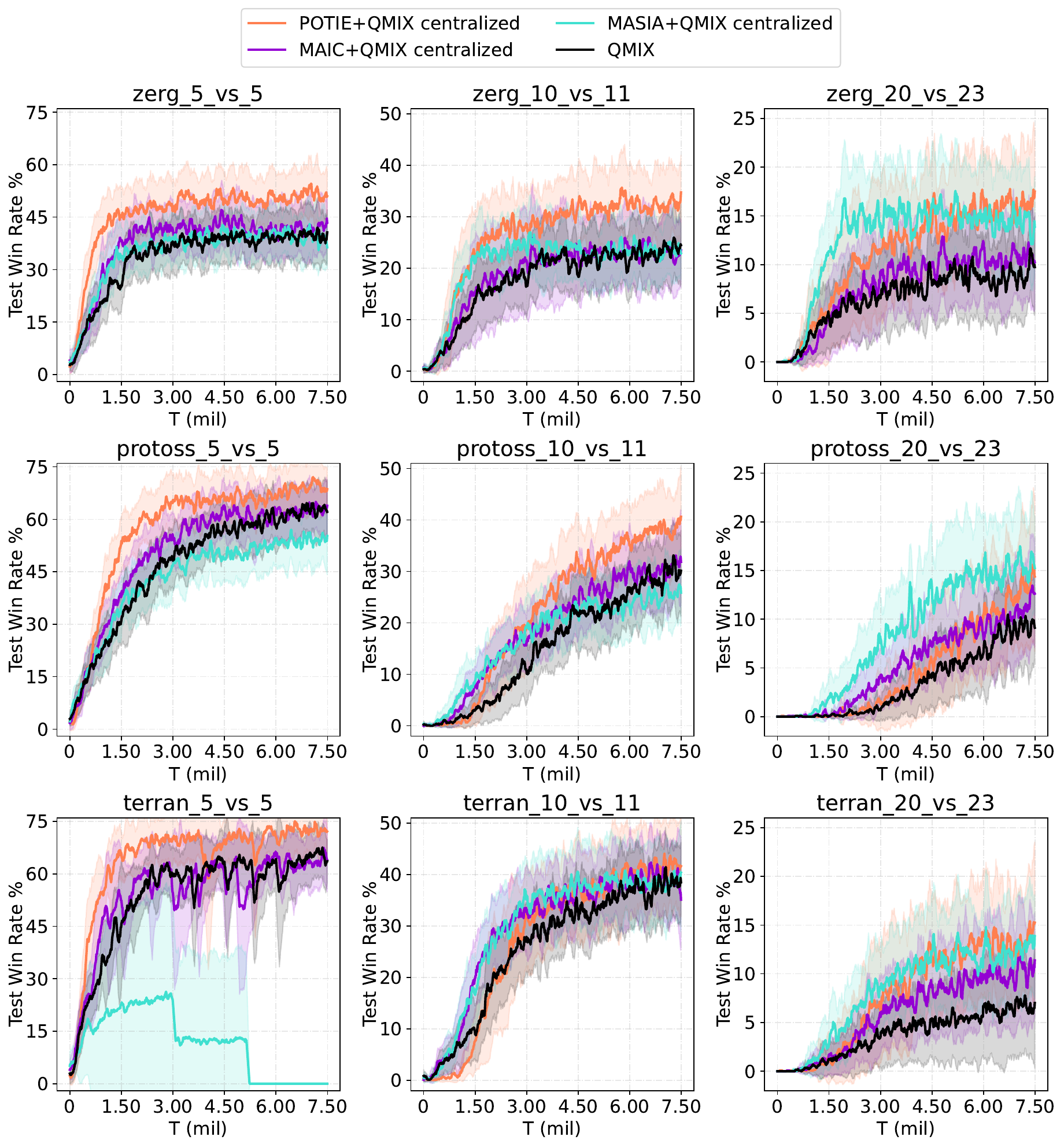}
  \caption{Comparison of different communication algorithms with the QMIX mixer on a variety of SMACv2 maps. \ourcom{} exhibits robust performance. Even though MASIA performs better on some maps, it performs worse than the bare mixer on others.}
  \label{fig:POTIE_vs_comm_qmix}
\end{figure*}

\begin{figure*}
  \centering
  \includegraphics[width=0.79\textwidth]{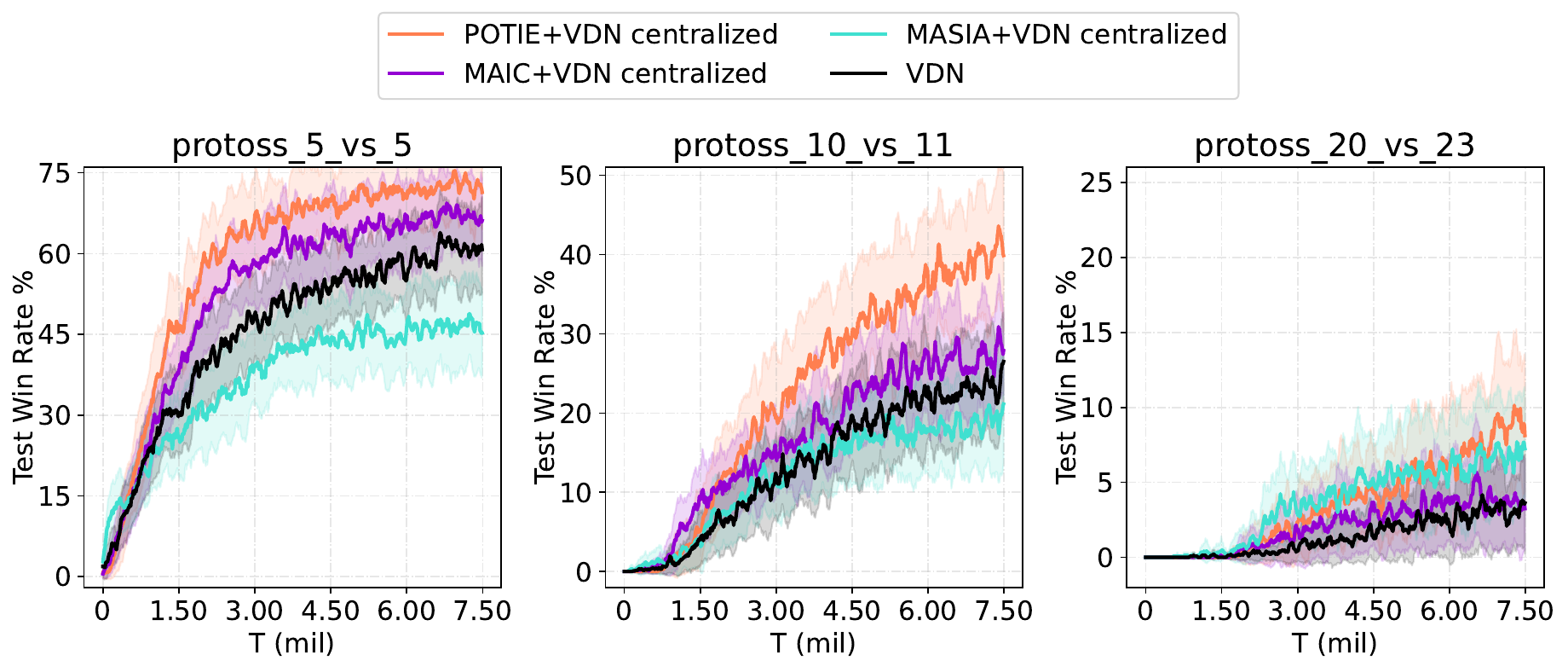}
  \caption{Comparison of different communication algorithms with the VDN mixer. \ourcom{} shows good performance with this mixer.}
  \label{fig:POTIE_vs_comm_vdn}
\end{figure*}

\begin{figure*}
  \centering
  \includegraphics[width=0.79\textwidth]{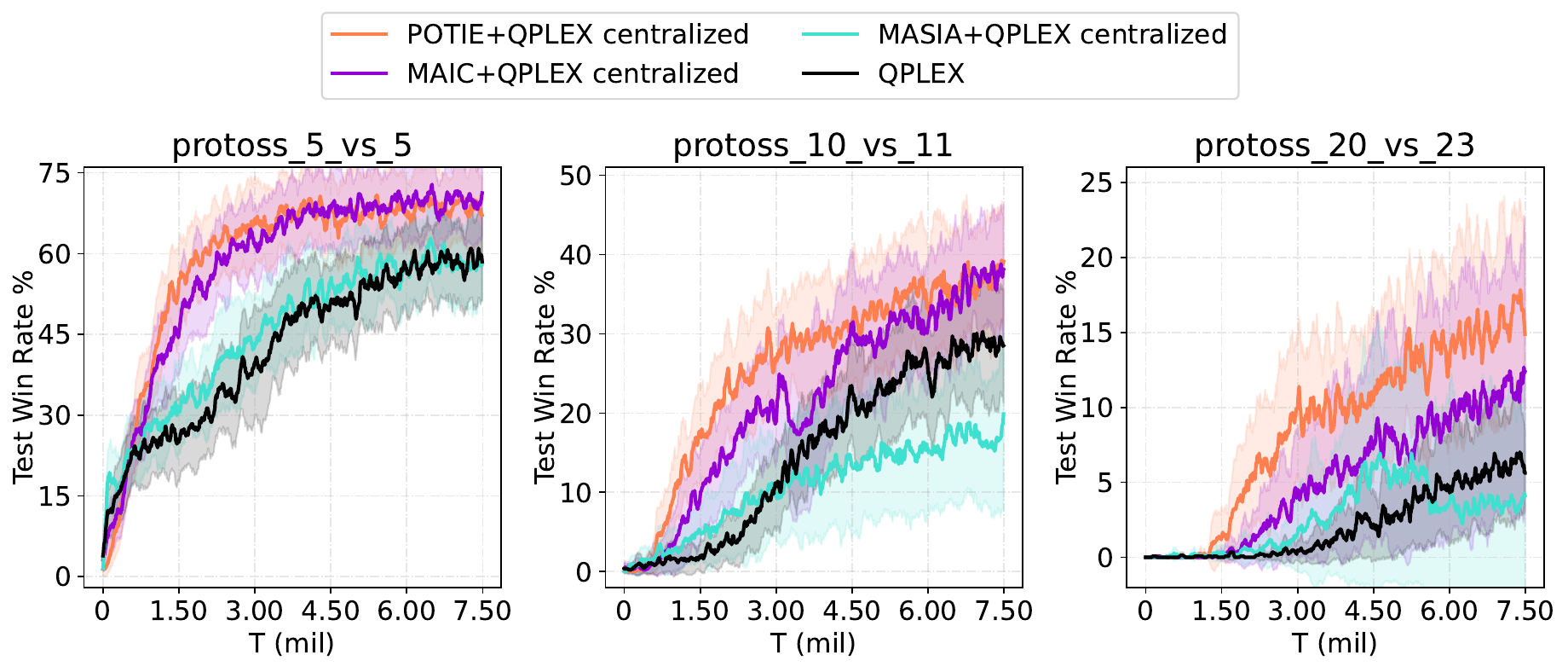}
  \caption{Comparison of different communication algorithms with the QPLEX mixer. \ourcom{} shows good performance with this mixer.}
  \label{fig:POTIE_vs_comm_qplex}
\end{figure*}

This section compares the performance of \ourcom{} with that of comparable MARL communication algorithms. Fig. \ref{fig:POTIE_vs_comm_qmix_smacv1} and Fig. \ref{fig:POTIE_vs_comm_qmix} present good performance in most SMAC and SMACv2 scenarios. Moreover, Fig. \ref{fig:POTIE_vs_comm_vdn} and Fig. \ref{fig:POTIE_vs_comm_qplex} show that \ourcom{} achieves good performance when combined with other mixers, namely VDN and QPLEX, on challenging SMACv2 maps.

\subsection{Hallway study}

\begin{figure}
  \centering
  \includegraphics[width=0.44\textwidth]{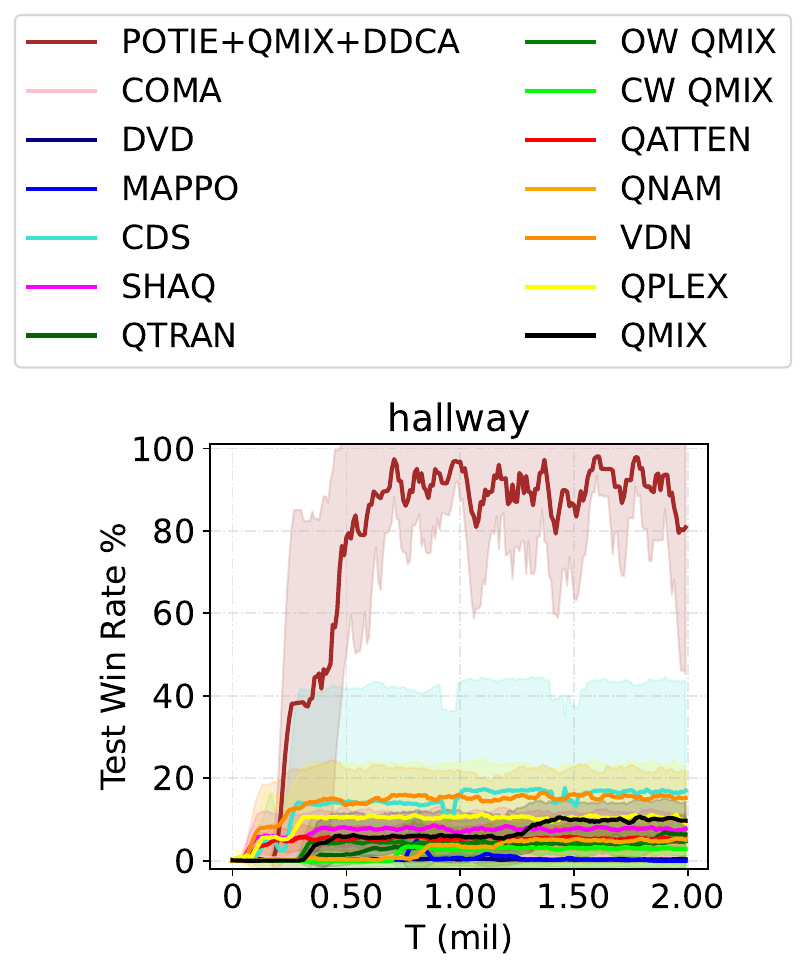}
  \caption{Evaluation of the CTDE baselines on Hallway. We observe that none solves this task.}
  \label{fig:app_Hallway_CTDE}
\end{figure}

\begin{figure}
  \centering
  \includegraphics[width=0.4\textwidth]{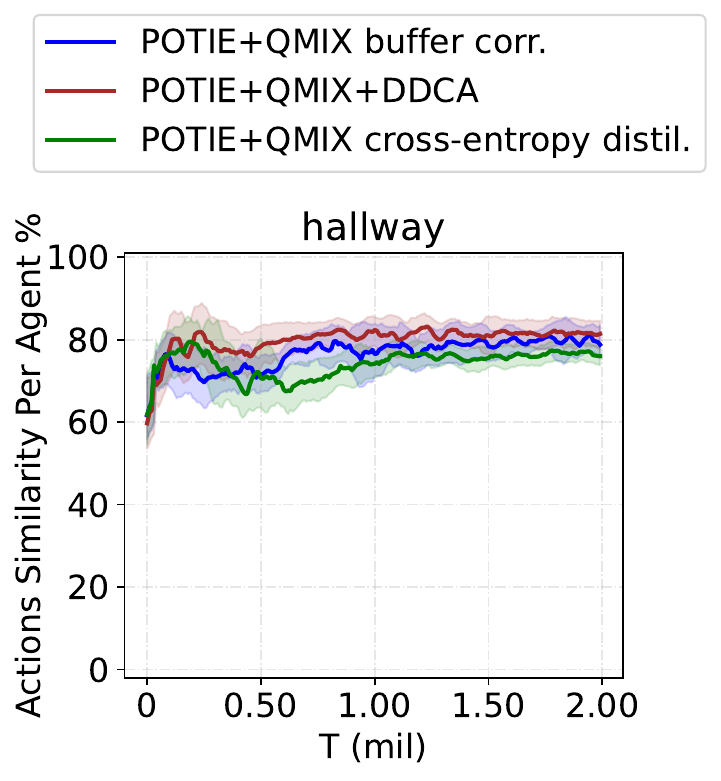}
  \caption{Comparison of the percentage of agents' actions chosen consistently between the centralized teacher and the decentralized student. We observe that \ourall{} reduces the imitation gap the most.}
  \label{fig:app_Hallway_actions_per_agent}
\end{figure}

\begin{figure}
  \centering
  \includegraphics[width=0.39\textwidth]{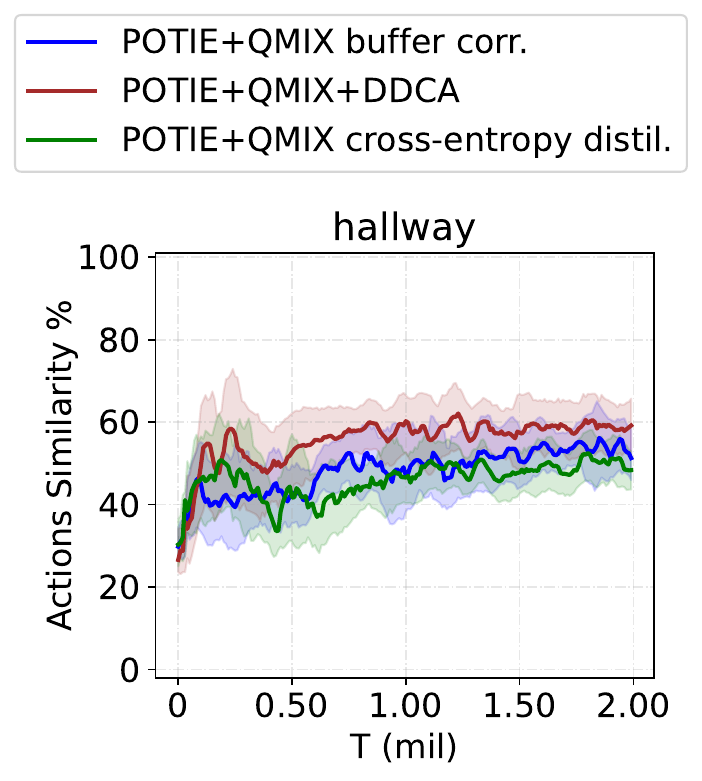}
  \caption {Comparison of the percentage of the joint actions chosen consistently between the centralized teacher and the decentralized student. We observe that \ourall{} reduces the imitation gap the most.}
  \label{fig:app_Hallway_actions}
\end{figure}

We further study the Hallway benchmark. First, we evaluate a broad range of CTDE algorithms to verify that none of them solves this task. The results are presented in Fig. \ref{fig:app_Hallway_CTDE}. We evaluated VDN \citep{2018sunehag+10}, QMIX \citep{2018rashid+5}, two variants of Weighted QMIX \citep{2020rashid+3}, QPLEX \citep{2021wang+4}, Qatten \citep{2020yang+6}, QTRAN \citep{2019son+4}, NA$^2$Q \citep{2023liu+2}, MAPPO \citep{yu2022the}, COMA \citep{coma}, DVD \citep{2022li+7}, SHAQ \citep{2022wang+3}, and CDS \citep{2021li+5}. None of them solved the task.

We study the reduction in imitation gap achieved by the corrections applied in \ourall{}. To measure this, we log the percentage of actions that were performed consistently between the centralized teacher and the decentralized student. We log this metric per agent and the whole team of agents. The results are presented in Fig. \ref{fig:app_Hallway_actions_per_agent} (per agent metric) and Fig. \ref{fig:app_Hallway_actions} (per team metric). We observe that our corrections reduce the imitation gap.

\subsection{LBF benchmark}

\begin{figure}
\centering
 \includegraphics[width=0.36\textwidth]{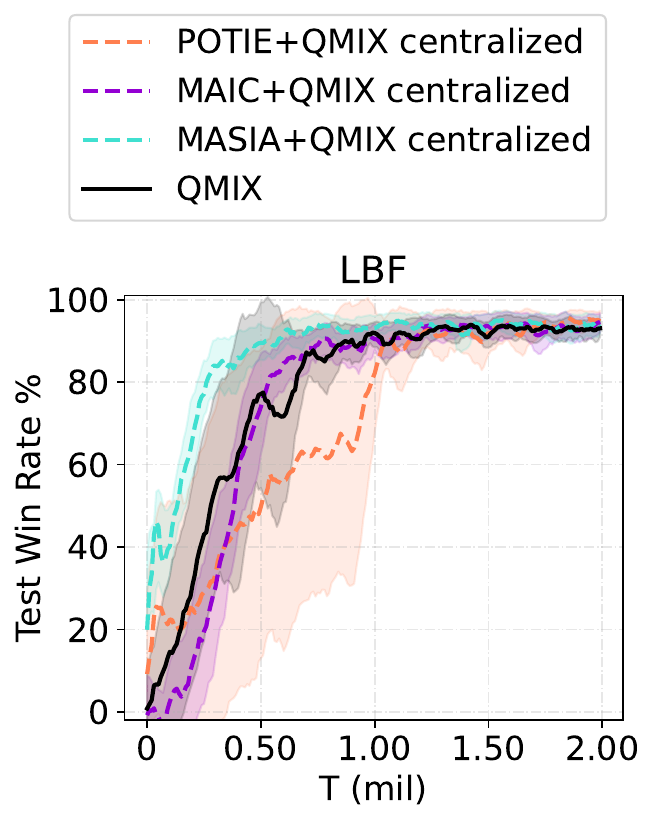}
  \caption{Comparison of \ourcom, MAIC, MASIA, and the bare mixer on LBF. All methods are successful in this task, although \ourcom{} learns more slowly.}
  \label{fig:MACTAS_comm_easyenv}
\end{figure}

\begin{figure}
  \centering \includegraphics[width=0.39\textwidth]{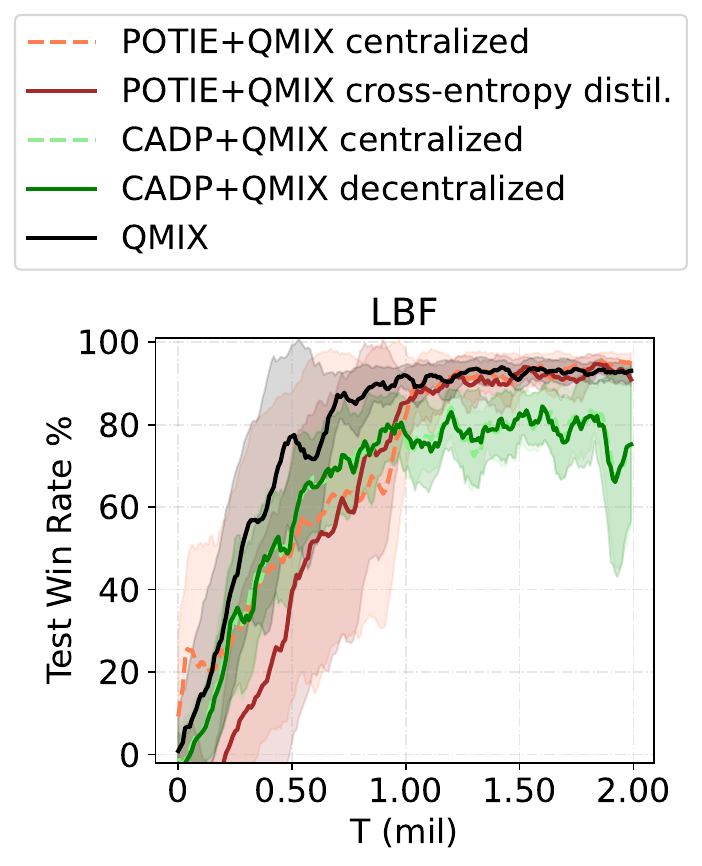}
  \caption{Comparison of the cross-entropy-based distillation, CADP, and the bare mixer. Cross-entropy-based distillation is enough to match the performance of the centralized counterpart.}
  \label{fig:MACTAS_vs_CADP_easyenv}
\end{figure}

LBF \citep{papoudakis2021foraging} is a grid world (we use $10 \times 10$) with agents and food. Each agent and each food item has a level, which is randomly assigned (in our experiments, the agent's level is at most 3 and the food item's level is at most 2). An agent can move in four directions, perform the "none" action, and load food. To collect the food, a group of agents must perform the load action together, and the sum of their levels must be at least as great as the food item's level. Agents receive a reward for eating food, and their goal is to maximize the return over a fixed horizon (in our experiments, set to 50 steps). The maximal return is normalized to 1. Moreover, each agent can observe only within a fixed radius around itself (we use a~radius of 2).

We report training curves for \ourcom{} against centralized baselines on the LBF task in Fig.~\ref{fig:MACTAS_comm_easyenv} and for cross-entropy distillation against CADP in Fig. \ref{fig:MACTAS_vs_CADP_easyenv}.

\subsection{Quartile-based evaluation}
\begin{figure*}
  \centering
  \includegraphics[width=0.99\textwidth]{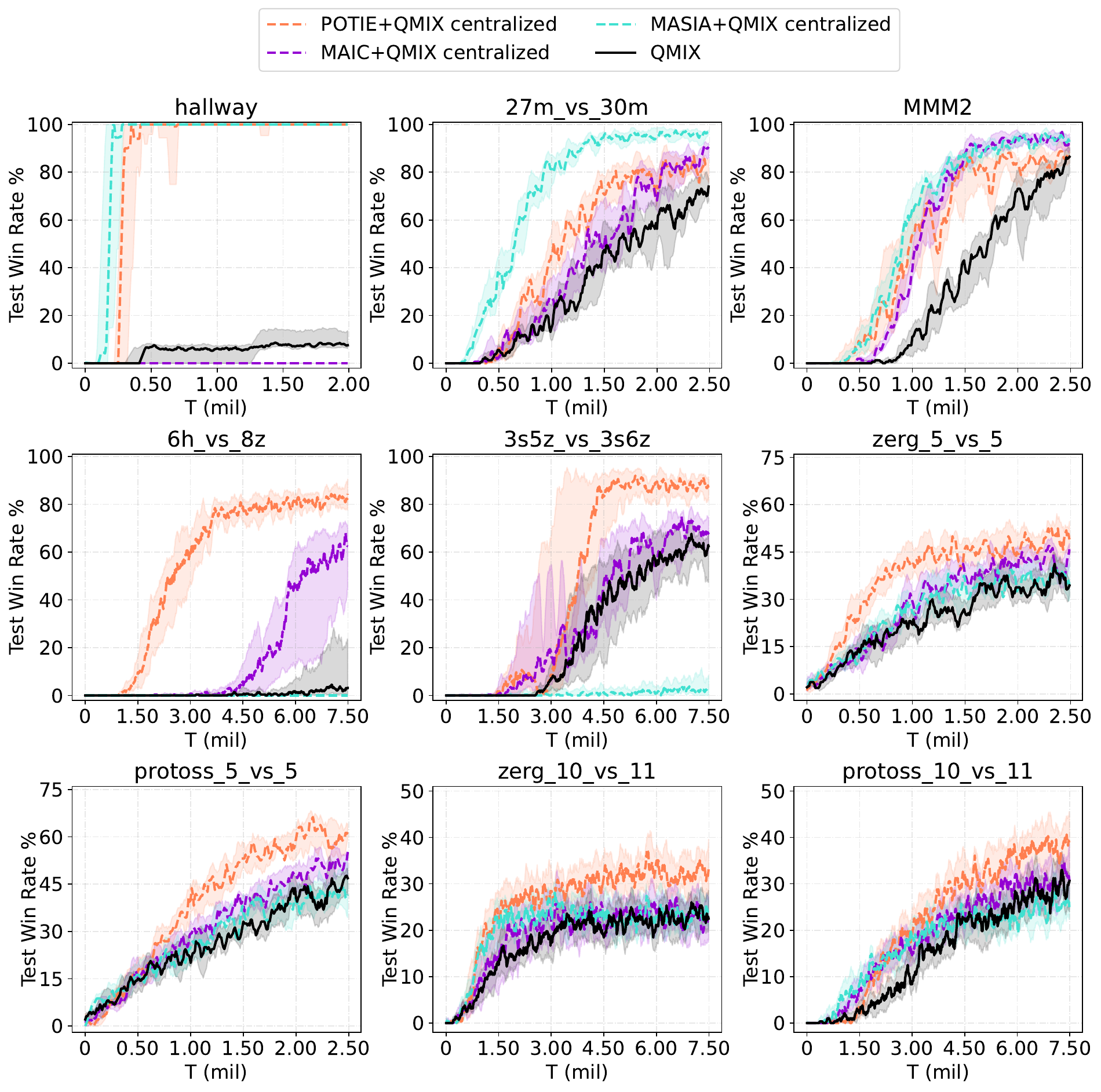}
  \caption{Comparison of \ourcom{}, MASIA, MAIC, and bare QMIX with interquartile ranges. \ourcom{} achieves the best results on all SMACv2 and the hardest SMAC maps.}
  \label{fig:app_POTIE_IQM}
\end{figure*}

\begin{figure*}
  \centering
  \includegraphics[width=0.6\textwidth]{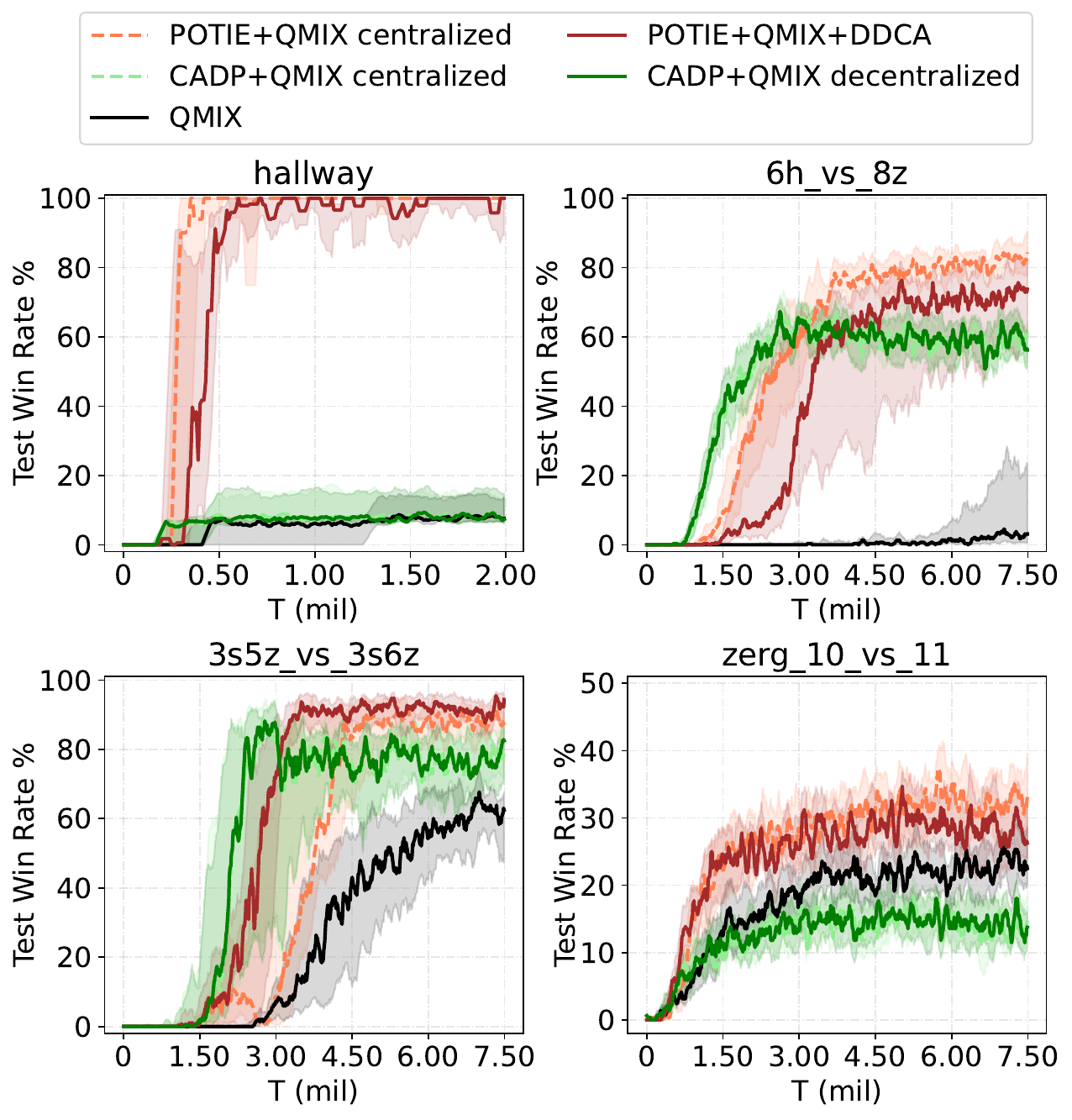}
  \caption{Comparison of \ourall{}, CADP, and bare QMIX with interquartile ranges. \ourcom{} achieves the best results.}
  \label{fig:app_DDCA_IQM}
\end{figure*}

To investigate the impact of outlier observations, we present the main results from the article with a quartile-based metric. Figures~\ref{fig:app_POTIE_IQM} and~\ref{fig:app_DDCA_IQM} show the IQM as a central line, with the $1^{st}$-to-$3^{rd}$-quartile range indicated by the shaded area.

\section{Additional ablations}
\label{sec:ablations_extra}

In this section, we present additional ablation studies.

\subsection{Robustness study}

\begin{figure*}
  \centering
  \includegraphics[width=0.89\textwidth]{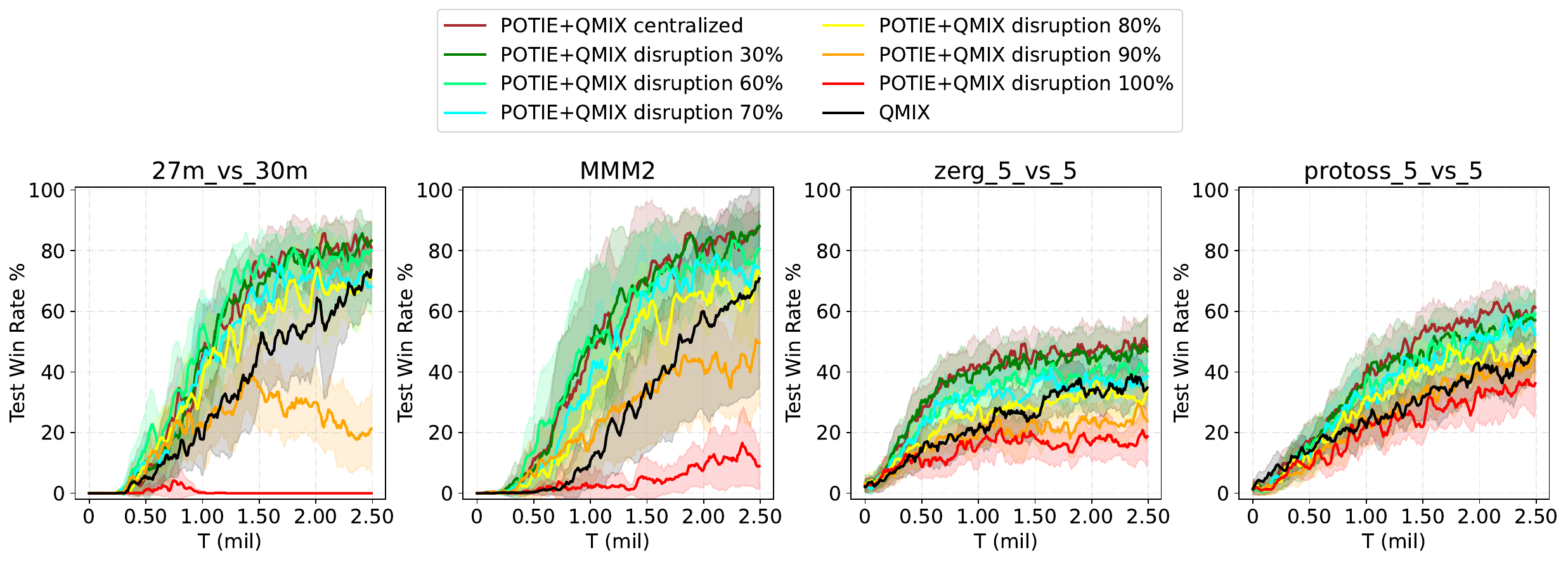}
  \caption{Results of connectivity-disruption simulations for different values of $p$, evaluated in test games during the training of \ourcom{} with QMIX on the SMAC and SMACv2 benchmarks. We observe robustness to communication-disruption rates of up to 60\%.}
  \label{fig:app_comm_disruptions}
\end{figure*}

Because RL algorithms are studied in synthetic environments, many of the challenges encountered in real-world applications are not adequately captured in these environments. In MARL, algorithms are typically designed to maximize learning speed across a range of predefined environments, with both the environment and the algorithm code executed on a single machine. This setup can be difficult to replicate in many real-world scenarios, where each agent is physically autonomous and performs all computation on embedded hardware, while communication with its peers must be conducted over radio, subject to volatility and disruptions.

We consider robustness an important aspect of algorithm design, and thus test \ourcom{} under such adverse conditions. The exact artifacts that may arise from communication channel instability vary with the communication medium and protocol, but, assuming proper error detection, they all result in the agents' inability to exchange messages for prolonged periods. To simulate this, we perturb the Transformer source mask, thereby preventing agent pairs from exchanging information over a sequence of time steps.
We validate \ourcom{} in these conditions by simulating network volatility -- the connectivity state $c_{i,j}$ between agents $i$ and $j$ at time $t$ is updated as follows:
\begin{equation}
c_{i,j}^{(t)} = 
\begin{cases} 
X \sim \text{Bernoulli}(\overline{p}) & \text{with probability } q \\
c_{i,j}^{(t-1)} & \text{with probability } 1-q 
\end{cases}
\end{equation}
where $q$ is the state-change probability (we use 10\%) and $\overline{p}$ is the probability of a functional connection (\% disruption).

We do not modify the \ourcom{} learning algorithm, keeping it unaware of the disruptions. The attention masking matrix is not stored in the state and is not provided to the model as input during inference or training.
Fig.~\ref{fig:app_comm_disruptions} presents the results of various disruption probabilities $p$ across SMAC and SMACv2 environments. \ourcom{} performs consistently, demonstrating strong performance under disruption rates of up to 60\%.

The results indicate that, without any changes to the algorithm, \ourcom{} could be deployed under a temporally varying loss of up to $60\%$ of pairwise connections, and it shows strong learning performance.

\section{Hyperparameters}
\label{sec:hyperparams}

\begin{table*}
\centering
\begin{tabular}{lccc}
\hline
\textbf{Scenario} & \!\!\!\!\!\!\!\!\!\!\!\!\!\!\!\!\textbf{Stacked encoders } & \textbf{FFN dimension} & \textbf{Difficulty} \\ \hline
6h\_vs\_8z, 3s5z\_vs\_3s6z, all SMACv2 maps\!\!\!\!\!\!\!\!\!\!\!\!\!\!\!\! & 3             & 512 & Hard                      \\
2c\_vs\_64zg, 5m\_vs\_6m, 27m\_vs\_30m, MMM2    & 1             & 256 & Medium                   \\
LBF and Hallway                         & 1             & 64 & Easy
\end{tabular}
\caption{\ourcom{} architecture for each scenario.}
\label{tab:shapes}
\end{table*}

\begin{figure*}
  \centering
  \includegraphics[width=0.89\textwidth]{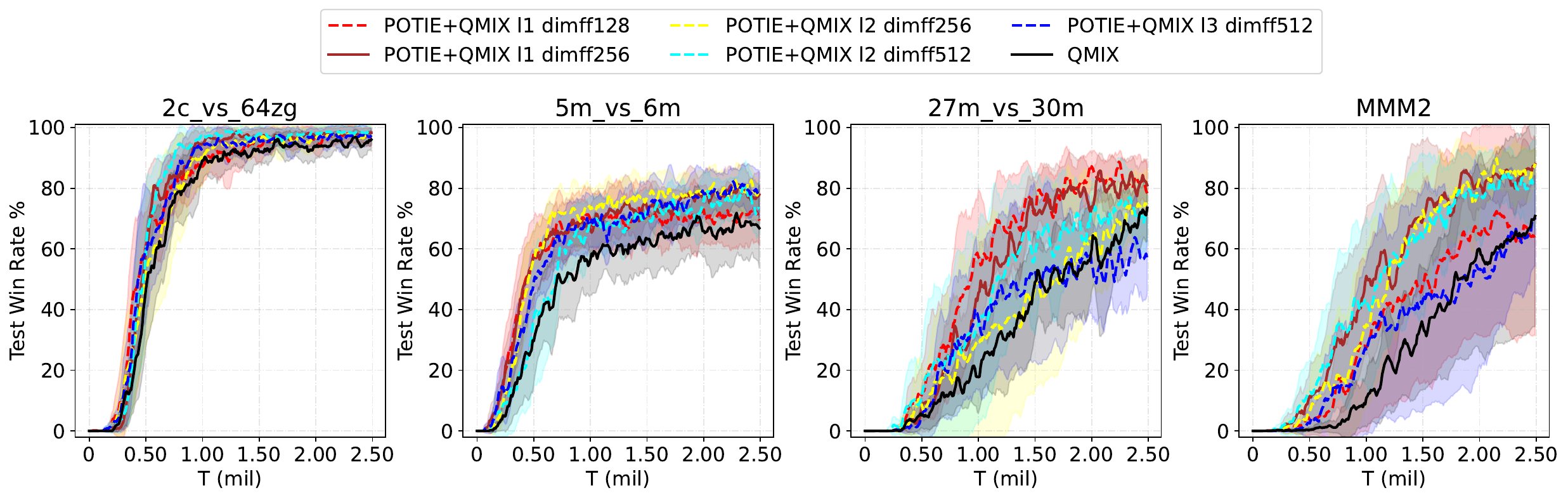}
  \caption{Comparison of different neural network sizes for the \ourcom{} algorithm on the simple SMAC scenarios. We observe that two environments - MMM2 and 27m\_vs\_30m are sensitive to the architectural choice.}
  \label{fig:POTIE_arch_grid_smacv1}

\centering
  \includegraphics[width=0.49\textwidth]{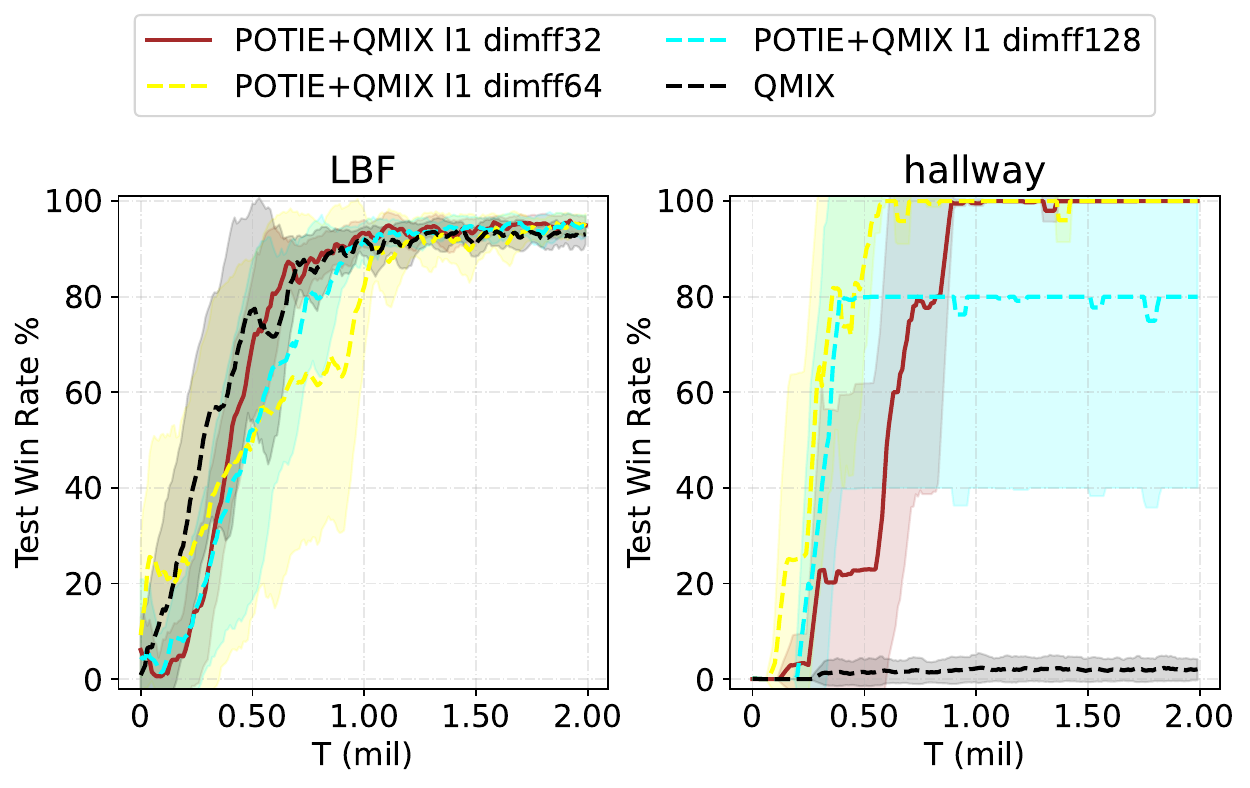}
  \caption{Comparison of different neural network architectures for the \ourcom{} algorithm on the easy environments. We observe that even a slightly larger feedforward network dimension may worsen performance on LBF or Hallway.}
  \label{fig:app_POTIE_arch_grid_easyenv}

\centering
  \includegraphics[width=0.49\textwidth]{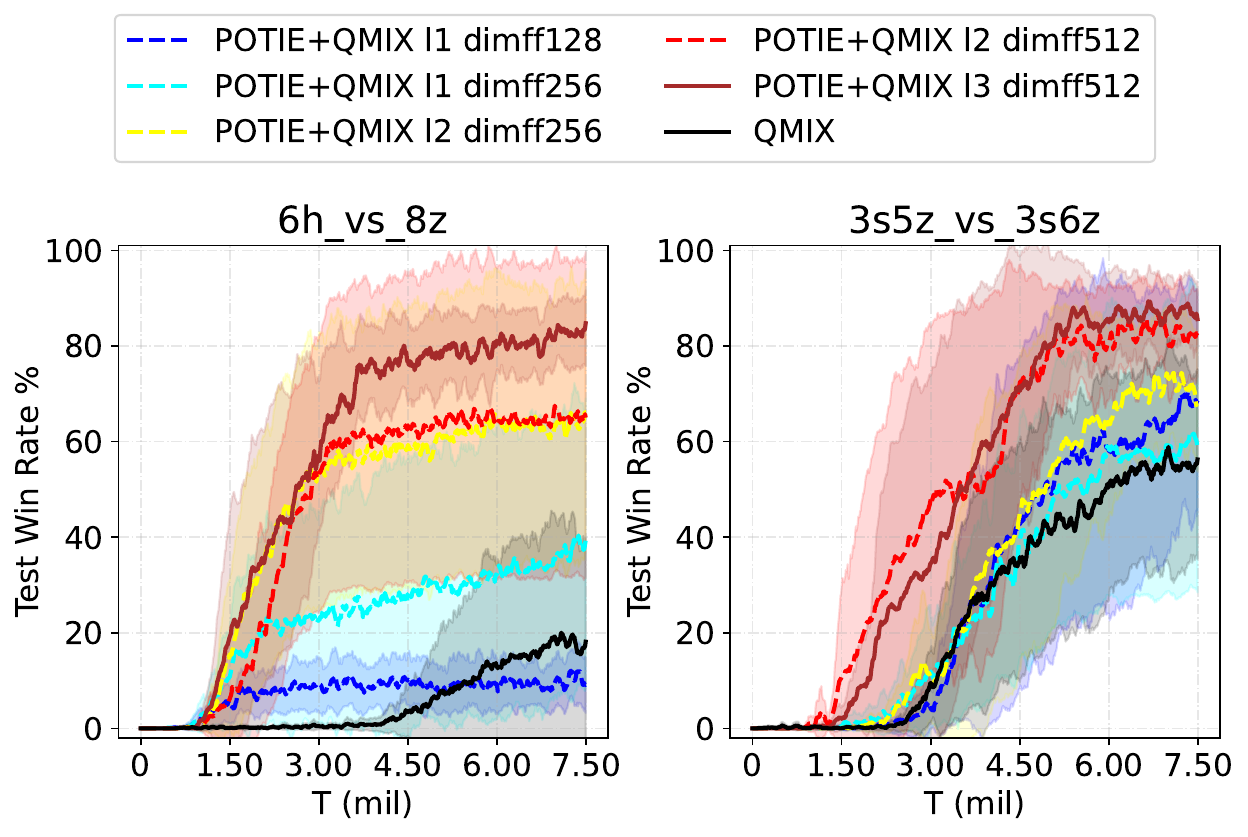}
  \caption{Comparison of different neural network architectures for the \ourcom{} algorithm on the hardest SMAC maps. We observe that larger architectures perform better.}
  \label{fig:POTIE_arch_grid_smacv1_big}
\end{figure*}

\begin{figure*}
  \centering
  \includegraphics[width=0.89\textwidth]{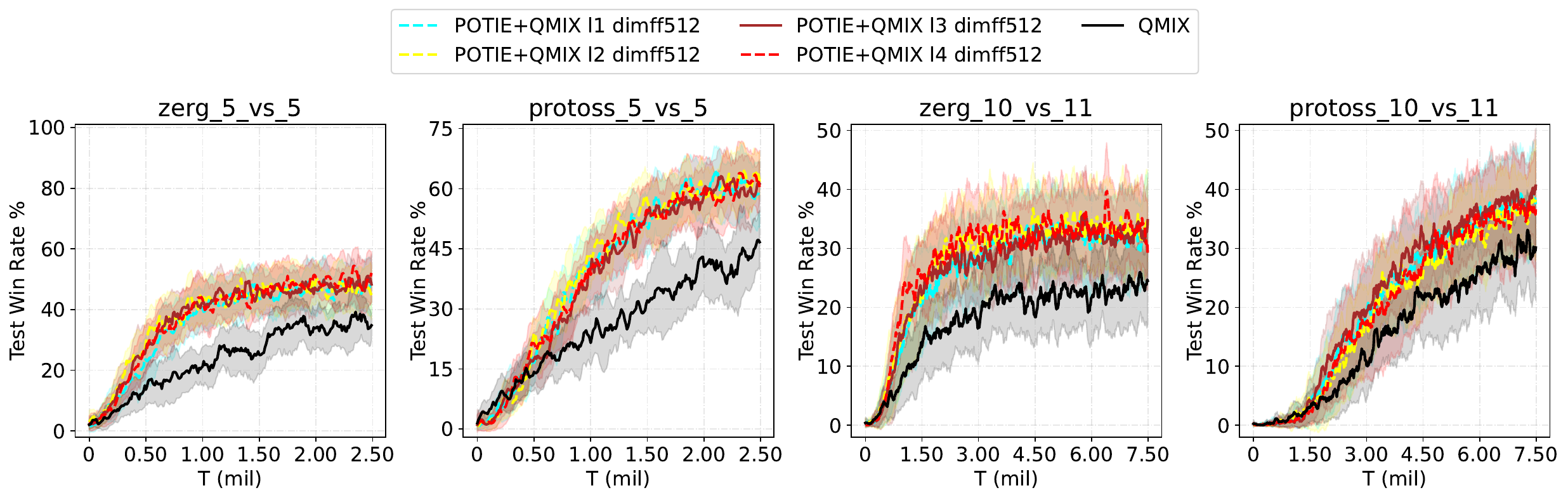}
  \caption{Comparison of different numbers of layers in the \ourcom{} algorithm on SMACv2 scenarios. We obtain similar results for different numbers of Transformer layers.}
  \label{fig:POTIE_arch_grid_smacv2}
\end{figure*}

To assess the reliability of the algorithms' results, we investigate the impact of hyperparameter changes on \ourcom{} and CADP in this section.

\subsection{\ourcom{} hyperparameter study}

The architectures chosen in the main section of this work are listed in Table \ref{tab:shapes}.

The simplest environments require a smaller \ourcom{} architecture than the more complex environments do. The analysis is presented in Fig.~\ref{fig:app_POTIE_arch_grid_easyenv}. The SMAC maps require some tuning of the \ourcom{} communication channel size to achieve optimal performance, although almost all settings outperform the bare mixer. The grid is shown in Fig.~\ref{fig:POTIE_arch_grid_smacv1} and Fig.~\ref{fig:POTIE_arch_grid_smacv1_big}. The most challenging SMACv2 scenarios, shown in Fig.~\ref{fig:POTIE_arch_grid_smacv2}, exhibit a minor dependence on network size. This indicates robust performance on hard problems and randomized episode starting points, as well as the ability to generalize.
The variance in the results stems from the fact that all SMAC maps use a fixed initial episode state, which requires reproducible strategies.

\begin{figure}
  \centering
  \includegraphics[width=0.425\textwidth]{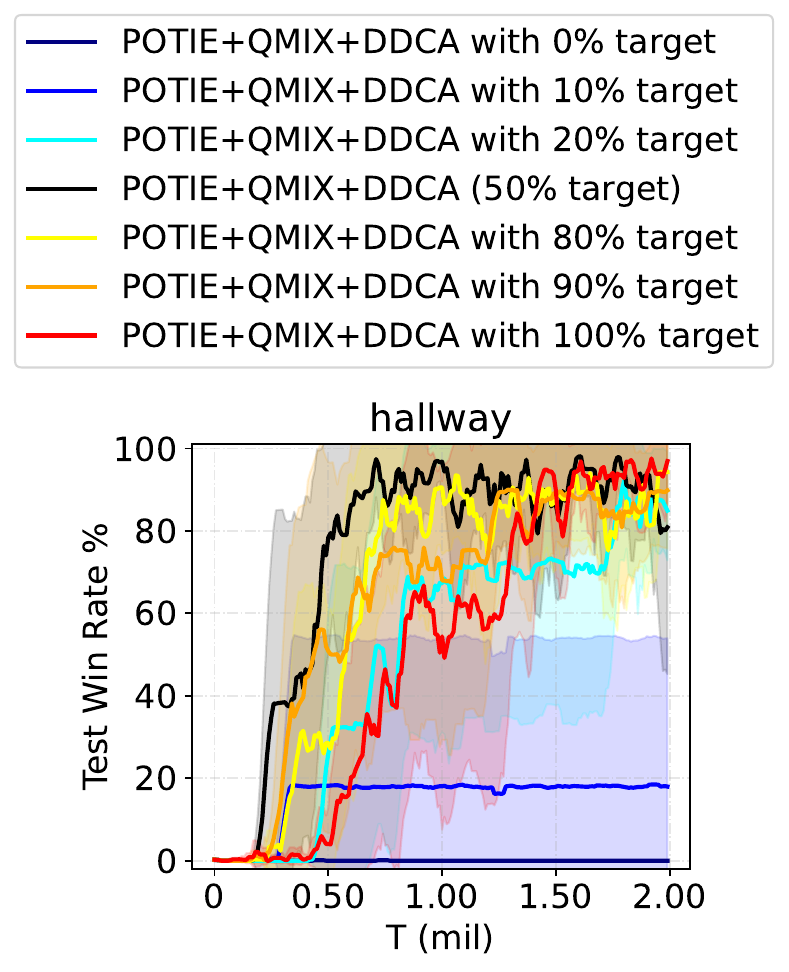}
  \caption{The analysis of the target correction for \ourall{}. We observe that using a probability of $20\%$ or higher for choosing the target action from the centralized teacher allows us to solve the Hallway task. The best performance is achieved for $50\%$.}
  \label{fig:app_Hallway_grid_target}
\end{figure}

\begin{figure}
  \centering
  \includegraphics[width=0.455\textwidth]{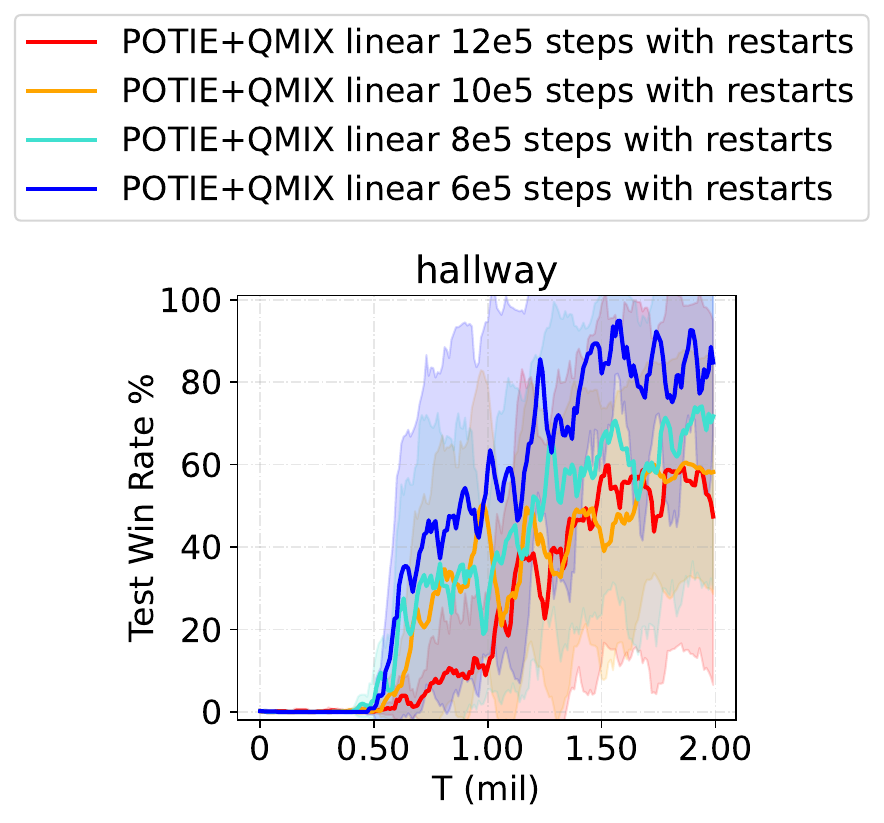}
  \caption{The analysis of linear scheduling with restarts. In this approach, annealing is performed during the whole training; the steps hyperparameter tells us after how many steps we get from 1 to 0. Then we restart the scheduling - we start again with 1 and schedule to 0. We observe that this method is a good alternative to exponential scheduling.}
  \label{fig:app_Hallway_grid_linear_restarts}
\end{figure}

\begin{figure*}
  \centering
  \includegraphics[width=0.69\textwidth]{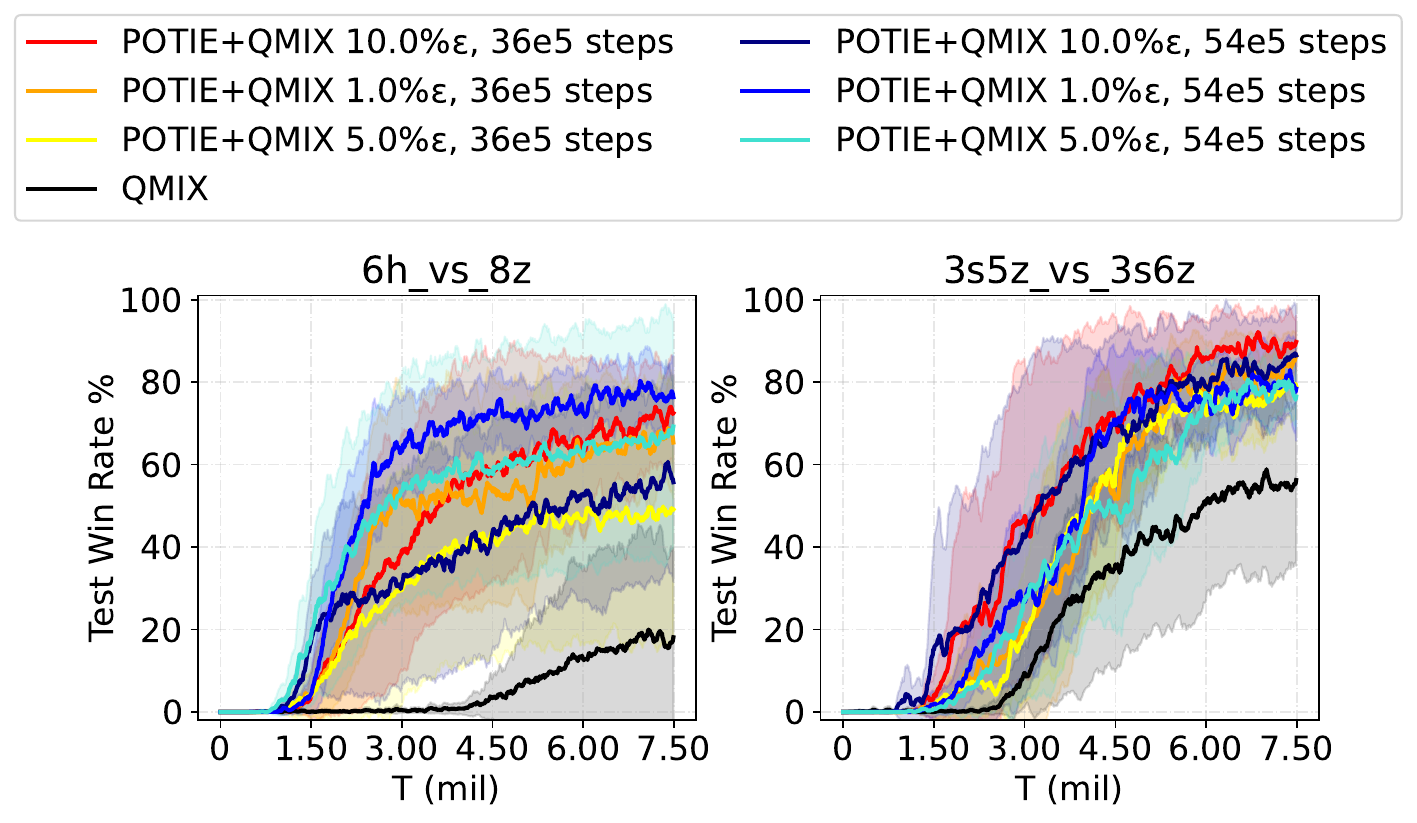}
  \caption{The analysis of the exponential annealing parameters for the replay buffer correction on the StarCraft-based environments. We observe that different settings yield satisfactory results as well.}
  \label{fig:app_grid_smac}
\end{figure*}

\begin{figure}
  \centering
  \includegraphics[width=0.37\textwidth]{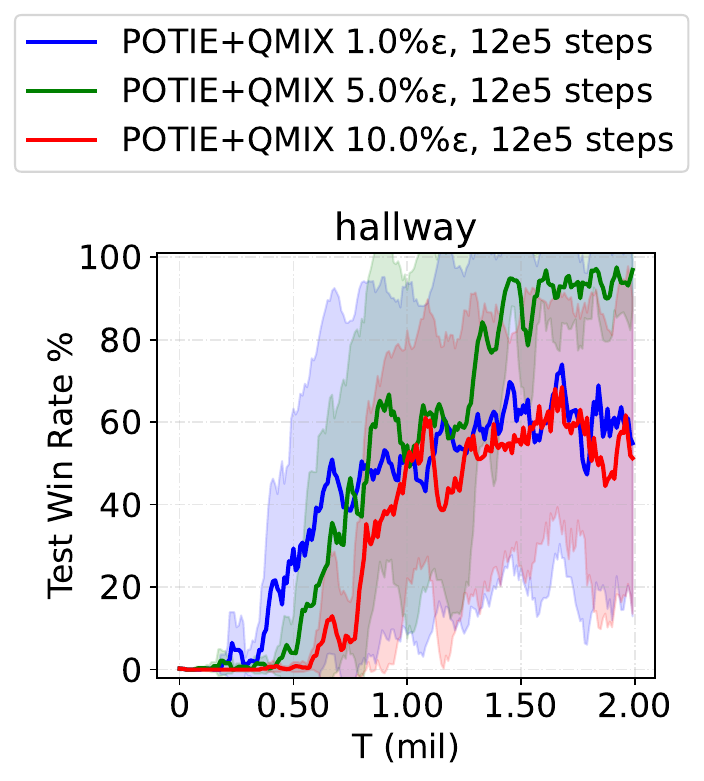}
  \caption{Grid search over the $\varepsilon$ parameter for the replay buffer correction with annealing time fixed at 1,200,000 steps. Suboptimal solutions solve Hallway in 3 out of 5 cases.}
  \label{fig:app_Hallway_grid_12e5}
\end{figure}

\begin{figure}
  \centering
  \includegraphics[width=0.365\textwidth]{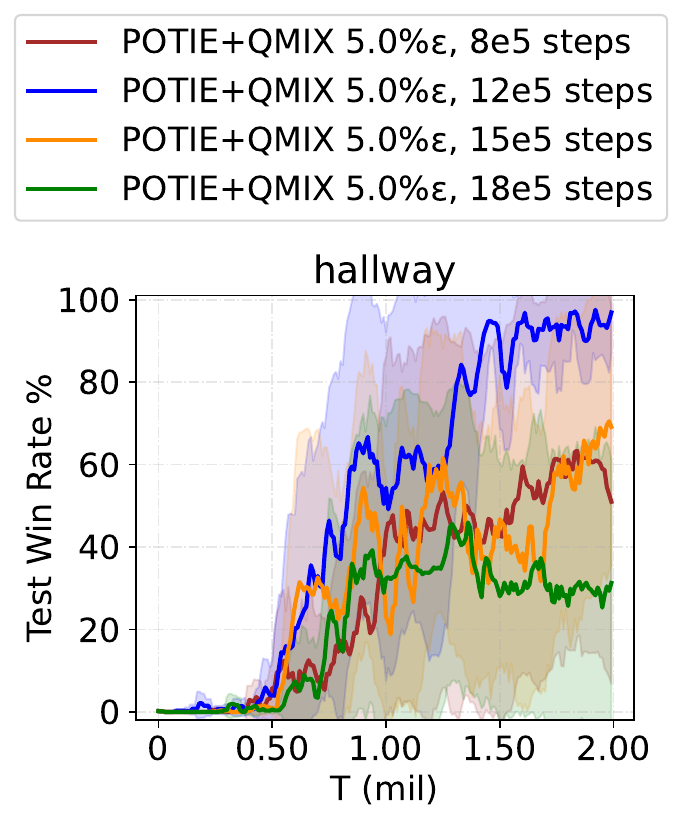}
  \caption{Grid search over the annealing time parameter for the replay buffer correction with $\varepsilon$ fixed at $5\%$. Two suboptimal solutions solve Hallway in 3 out of 5 cases.}
  \label{fig:app_Hallway_grid_epsilon}
\end{figure}

\begin{figure}
  \centering
  \includegraphics[width=0.38\textwidth]{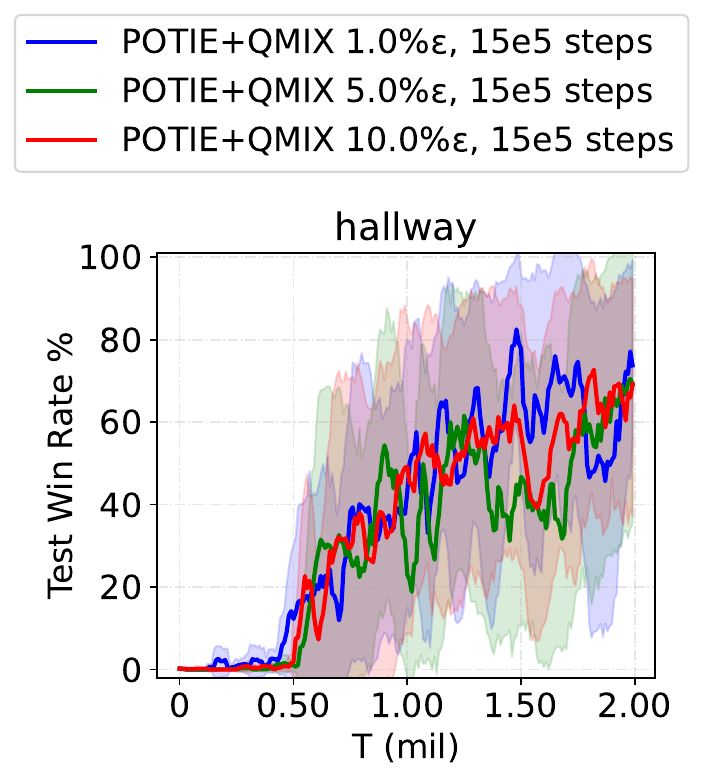}
  \caption{Grid search over the $\varepsilon$ parameter for the replay buffer correction with annealing time fixed at 1,500,000 steps. The setup used in StarCraft-based environments solves Hallway in 4 out of 5 cases.}
  \label{fig:app_Hallway_grid_15e5}
\end{figure}

We tested a piecewise linear scheduling instead of exponential scheduling for the replay buffer correction in \ourall{}. We schedule linearly from 1 to 0 and then restart to 1 and keep repeating the procedure. We present the results of a small grid search of this approach on Hallway in Fig. \ref{fig:app_Hallway_grid_linear_restarts}. We observe that it is a good alternative to exponential scheduling.

For the replay buffer correction in \ourall{}, we performed a small grid search over the hyperparameters $\epsilon_s$ and $time$ of the exponential scheduling. We also performed a small grid search over the hyperparameter of the target imitation gap correction on the Hallway task; the results are presented in Fig. \ref{fig:app_Hallway_grid_target}. For the StarCraft-based environments, we chose $time$ equal to $72\%$ of the training time and $\epsilon_s = 1\%$; however, different settings also yield satisfactory results. The study of the grid search over StarCraft-based environments is presented in Fig. \ref{fig:app_grid_smac}. We observed that to obtain an almost perfect score on Hallway, slightly different settings are required: $time$ equal to $60\%$ of the training time and $\epsilon_s = 5\%$. 
However, many other hyperparameter settings solve Hallway in at least $\frac{3}{5}$ of the cases; we present the study in Fig. \ref{fig:app_Hallway_grid_12e5}-\ref{fig:app_Hallway_grid_15e5}.

\subsection{CADP breakpoint parameter}

\begin{figure}
  \centering
  \includegraphics[width=0.49\textwidth]{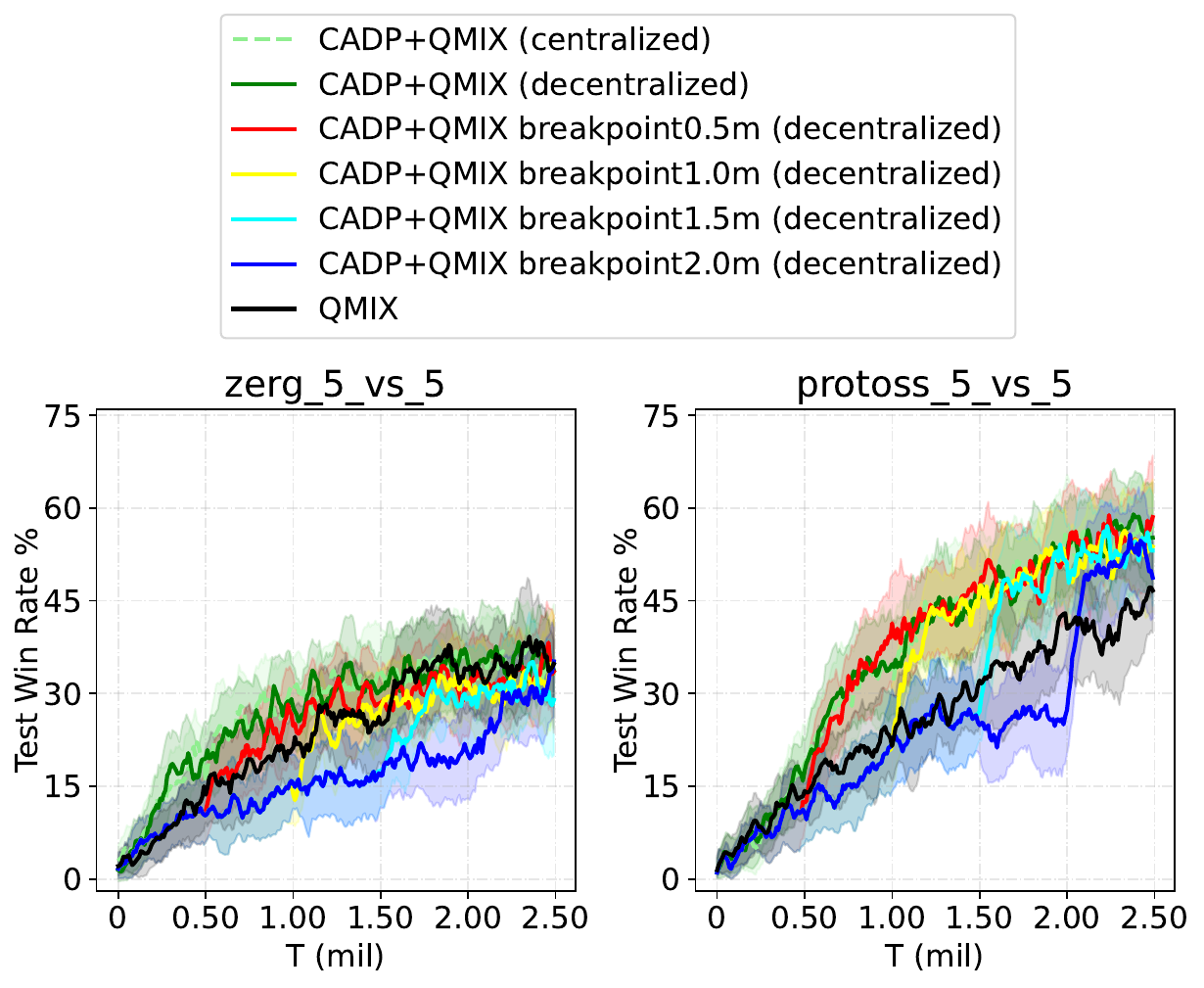}
  \caption{Grid search over the breakpoint parameter for the CADP algorithm. We observe that a value of 0 yields the best results.}
  \label{fig:CADP_breakpoint}
\end{figure}

We perform a grid search over the breakpoint parameter of the CADP method on $\verb|zerg_5_vs_5|$ and $\verb|protoss_5_vs_5|$ SMACv2 maps. The results are presented in Fig.~\ref{fig:CADP_breakpoint}. We observe that setting this parameter to zero yields the best performance during decentralized execution, consistent with the learning curves reported in the original article \citep{2025zhou+6CADP}.

\subsection{Setup and hyperparameters}
Each experiment is run for a fixed number of training episode steps -- 2.0 million for Level-Based Foraging \citep{papoudakis2021foraging} (results presented in the Appendix) and Hallway, 2.5 million for simpler SMAC and SMACv2 environments, and 7.5 million for the hardest SMAC and SMACv2 environments, interleaved with test episodes every 10000 steps. Each experiment is repeated for seeds $1,\dots, 5$. We report the smoothed average percentage of wins in test games, along with the standard deviations on the plots.
For MAIC and MASIA communication, we use the same hyperparameters as in the original papers.
For CADP, we tune the breakpoint hyperparameter on the protoss\_5\_vs\_5 and zerg\_5\_vs\_5 maps and use the value 0, which performs best (consistent with the training curves reported in the original paper). Grid search over the breakpoint hyperparameter is available in the Appendix.

All hyperparameters not strictly related to the communication algorithm are shared between maps, algorithms, and mixers and are presented in Table~\ref{tab:basic:settings}.

\begin{table}
\centering
\begin{tabular}{ll}
\hline
\multicolumn{1}{l}{\textbf{Hyperparameter}} & \textbf{Value} \\ \hline
Batch size                                   & 32             \\
Test episodes                                & 32             \\
Replay buffer size                           & 5000           \\
Discount factor                              & 0.99           \\
Start epsilon                                & 1.0            \\
Finish epsilon                               & 0.05           \\
Anneal steps                                 & 50,000${}^1$ or 100,000${}^2$        \\
RNN units                                    & 64             \\
Optimizer                                    & RMSProp        \\
Learning rate                                & 0.0005
\\ \hline
\end{tabular}
\caption{Basic experimental settings and hyperparameters. ${}^1$~for LBF, Hallway, and 2.5-million step SMAC maps. ${}^2$~for SMACv2 and 7.5 million-step SMAC maps.}
 \label{tab:basic:settings}
\end{table}

\begin{table}
    \centering
    \begin{tabular}{|l|c|}
            \hline
        \textbf{Library name} & \textbf{Version}  \\
        \hline
        Python &  3.9.21 or 3.9.22  \\
        numpy & 1.23.1   \\
        nvidia-cublas-cu12 & 12.4.5.8   \\
        nvidia-cuda-cupti-cu12 & 12.4.127   \\
        nvidia-cuda-nvrtc-cu12 & 12.4.127   \\
        nvidia-cuda-runtime-cu12 & 12.4.127   \\
        nvidia-cudnn-cu12 & 9.1.0.70   \\
        nvidia-cufft-cu12 & 11.2.1.3   \\
        nvidia-curand-cu12 & 10.3.5.147   \\
        nvidia-cusolver-cu12 & 11.6.1.9   \\
        nvidia-cusparse-cu12 & 12.3.1.170   \\
        nvidia-nccl-cu12 & 2.21.5   \\
        nvidia-nvjitlink-cu12 & 12.4.127   \\
        nvidia-nvtx-cu12 & 12.4.127   \\
        scipy & 1.13.1   \\
        torch & 2.5.1 \\
        \hline
    \end{tabular}
    \caption{Essential libraries in the programming environment.}
    \label{tab:software}
\end{table}

\begin{table*}
    \centering
    \begin{tabular}{|p{1.5cm}|p{1.5cm}|p{1.5cm}|p{2cm}|p{3cm}|p{1cm}|}
        \hline
        \textbf{Cluster name} & \textbf{GPU device} & \textbf{GPU driver version} & \textbf{CPU device} & \textbf{Operating system} & \textbf{Total RAM}  \\
        \hline
        \textbf{Cluster 1} &  NVIDIA A100 SXM4 80GB & 535.154.05 & AMD EPYC 7742 64-Core Processor & Linux-5.15.0-1059-nvidia-x86 \_64-with-glibc2.35 & 1024 GB \\
        \hline
        \textbf{Cluster 2} &  NVIDIA A100 SXM4 40GB & 570.133.20 & AMD EPYC 7742 64-Core Processor & Linux-5.14.0-503.26.1.el9 \_5.x86\_64-x86\_64-with-glibc2.34 & 1024 GB\\
        \hline
        \textbf{Cluster 3} &  NVIDIA A100 PCIe 80GB & 565.57.01 & AMD EPYC 7713 64-Core Processor & Linux-6.8.0-64-generic-x86 \_64-with-glibc2.39 & 2048 GB \\
        \hline
    \end{tabular}
    \caption{Specification of the computational clusters used in the experiments.}
    \label{tab:hardware}
\end{table*}

\subsection{LBF and Hallway}
To ensure reproducibility, we report the exact map names used for Level-based Foraging (LBF) and Hallway. For LBF, we use lbf-2s-10x10-4p-2f, and for Hallway, we use hallway-3a-1g.

\section{Software}

Table \ref{tab:software} contains a list of key programming libraries used during our experiments. The complete list of programming libraries is available in the official code repository of \ourall{} in the file \verb|requirements.txt|.

\section{Hardware}

Here we present the hardware used in our experiments. We utilized the NVIDIA Ampere GPU architecture across three computational clusters. Overall hardware utilization time is estimated at around 80,000 GPU-hours. To replicate the experiments described in the main section of this article, approximately 3,000 GPU-hours are required. Around 7,000 GPU-hours are required to replicate additional experiments described in the appendix. An experiment using five random seeds takes approximately 1--5 days, depending on the environment type and the number of agents. The specific computational cluster setup is presented in Table \ref{tab:hardware}.